\documentclass[11pt,a4paper]{article}

\usepackage[margin=1in]{geometry}
\usepackage{newtxtext,newtxmath}
\usepackage[numbers,sort&compress]{natbib}
\usepackage{graphicx}
\usepackage{array}
\usepackage{booktabs}
\usepackage{multirow}
\usepackage{float}          
\providecommand{\sidecapoff}{}
\newcommand{\phead}[1]{\paragraph{#1.}}
\usepackage{ragged2e}   
\newcolumntype{L}[1]{>{\RaggedRight\arraybackslash}p{#1}}  
\usepackage{amsmath}
\usepackage{xcolor}
\usepackage{microtype}
\usepackage[font=small,labelfont=bf]{caption}
\usepackage[hidelinks]{hyperref}
\usepackage{xurl}                 
\newcommand{\sysname}[1]{\texttt{#1}}
\newcommand{\repo}[1]{\texttt{#1}}
\newcommand{\hfns}{the author's Hugging Face namespace}
\newcommand{\code}[1]{\texttt{#1}}

\graphicspath{{../figures/}{figures/}}

\title{\bfseries Winning the Leaderboard, Losing the Distribution:\\
Diagnosing and Mitigating Acoustic Narrowing in Low-Resource ASR}

\author{Kadir G\"oksel G\"und\"uz\thanks{ORCID: 0009-0007-9120-5659.
Email: \texttt{gunduz25@itu.edu.tr}.}\\
\normalsize Department of Energy Science and Technology, Energy Institute\\
\normalsize Istanbul Technical University, Istanbul, Turkey}

\date{\normalsize July 2026\thanks{Preprint. Results are reported as of the 2026-07-21
cut-off; Section~\ref{sec:setup} gives the full experiment chronology and states which
components were fixed before which run. No number reported here is estimated.}}

\begin{document}
\maketitle

\begin{abstract}
\noindent
For a low-resource language, one public leaderboard is often the only shared measure of
progress, so optimizing toward it means optimizing toward a test distribution. Our low-rank
fine-tune of \sysname{whisper-large-v3-turbo} ranked first on MediaSpeech, Common Voice~17 and the five-set average of a Turkish community board, at 9.02\% word error on MediaSpeech against 11.93\% for the best unmodified model. On the same board's private split it lost to the \emph{unmodified}
\sysname{whisper-large-v3}. 

We treat that split as an alarm, not as evidence. The finding does not depend on it. On a
twenty-condition FLEURS-TR benchmark built by a group with no part in our model selection, the
unmodified base beats every one of our fine-tunes in every condition group, and all five
deficits clear zero under paired bootstrap. 
We call this failure mode \emph{acoustic narrowing}: fine-tuning contracts the acoustic range a
model serves well.

The most useful finding concerns measurement. The in-house proxy we built to detect narrowing
concealed it instead, ranking the narrowed fine-tune above the sturdier base. Model selection
therefore needs at least one proxy drawn from an \emph{independently produced}
distribution; a larger sample of the same source will not do. 

A rebuilt pipeline reduces the narrowing: a larger mixture with disclosed provenance, a
higher-capacity base, and base-referenced selection with trajectory averaging. It wins three
same-provenance held-out axes and draws level on FLEURS-TR, 8.16\% against 8.26\%, with the
paired interval spanning zero. That parity holds under a second inference stack, and training
the same recipe to a full epoch leaves the FLEURS-TR gap unchanged once the decoder is held
fixed. The recovery is one training trajectory, so we report it as a demonstration rather than
a validated recipe. 

Two experiments bound the claim. In Kazakh the inversion reappears on corpora we did not build.
With the same corpus and matched trainable-parameter shares, fine-tuning raised out-of-pool error by 2.23 points
for an attention encoder--decoder and lowered it by 3.46 for a CTC encoder-only model, each
across three seeds. Narrowing therefore depends on the system being adapted. 
We release TurkSpeak~Bench, a robustness-first Turkish ASR leaderboard. 

\end{abstract}

\vspace{0.5em}

\vspace{0.35em}
\noindent\textbf{Notation.} Three kinds of identifier appear in this paper, each in one fixed
form. \sysname{system names} are the models and evaluation sets we ran. \repo{repository names}
are Hugging Face repositories under \hfns, which is stated once under Data and code
availability and not repeated. \code{code tokens} are library and API names. \sysname{turbo} and \sysname{large-v3}
abbreviate \sysname{whisper-large-v3-turbo} and \sysname{whisper-large-v3} throughout. Signed
differences $\Delta$ are adapted minus base, so positive means worse, except where a figure
plots improvement and says so. Plain URLs appear only in that availability section and in the
title footnote.

\section{Introduction}
\label{sec:intro}

Public leaderboards now steer the development of automatic speech recognition (ASR). The Open
ASR Leaderboard fixes normalization, hardware and metrics, then ranks dozens of open and
proprietary systems~\citep{srivastav2025openasr}. It carries forward the philosophy of the ESB
benchmark. That benchmark warned that tuning to one dataset assumes knowledge of the test
distribution, and that such systems transfer poorly~\citep{gandhi2022esb}.

A leaderboard pulls hardest where there is nothing else to aim at. The Open ASR Leaderboard's
multilingual track covers a few European languages and no Turkish~\citep{srivastav2025openasr}.
For Turkish, the state of the art is defined by one community-run public benchmark on
Hugging Face, the ysdede Turkish ASR leaderboard~\citep{ysdede2026leaderboard}.

The accepted defence is evaluation data that submitters cannot see. Even the Open ASR
Leaderboard's maintainers list private held-out sets as future work~\citep{srivastav2025openasr}.
The Turkish board already has one. It caught us.

That split has two limits, and we state them at the outset. Its composition is not published,
so anything we say about its contents is inference. Only its maintainer can score a model on it,
so we could not re-run our recovery candidates there.

We therefore treat it as an \emph{alarm} rather than as evidence. No claim in this paper depends
on it. Everything we assert rests on evaluations a third party can audit and re-run: a
twenty-condition FLEURS-TR grid built by another team, the Kazakh corpora of
Section~\ref{sec:crosslingual}, and our own held-out axes, whose per-utterance outputs we
release. The same inversion appears on public data under controlled conditions, in a second
language, and in a controlled architecture comparison.

\begin{figure}[t]
\centering
\includegraphics[width=0.86\linewidth]{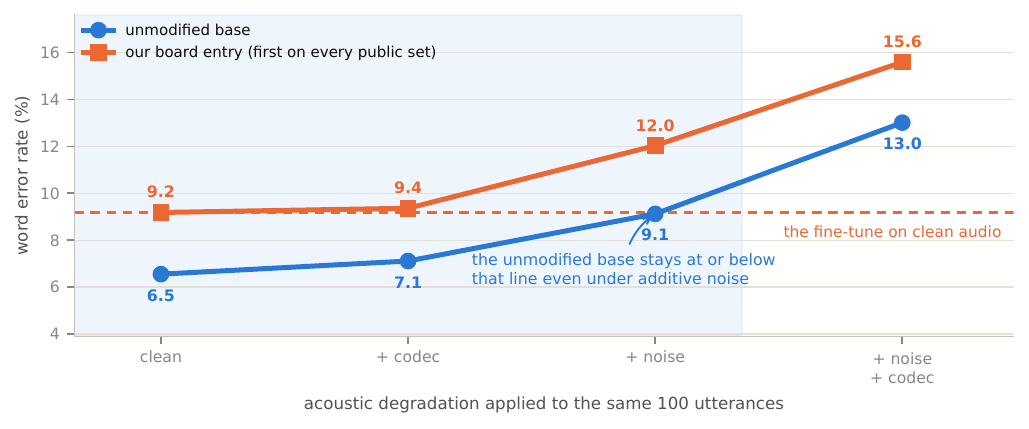}
\sidecapoff
\caption{Acoustic narrowing in one view. The same 100 FLEURS-TR utterances are scored under four
degradation severities by a team that did not build either model. Our board entry, which ranks
first on every public set its training corpus covered, is worse than the unmodified base at every
severity. The dashed line marks the entry's score on \emph{clean} audio, and the unmodified base
stays at or below it even under additive noise. What fine-tuning bought on the leaderboard it
paid for in the range of conditions the model still serves.}  
\label{fig:teaser}
\end{figure}

This paper reports the catch from the inside, and Figure~\ref{fig:teaser} states the result
before the argument.
We fine-tuned \sysname{whisper-large-v3-turbo}~\citep{radford2023whisper,openai2024turbo} with low-rank adaptation (LoRA)~\citep{hu2022lora}
on a 997.2\,h Turkish corpus assembled to track the board's public distributions: podcast, broadcast, lecture, and Common Voice speech. 
The leaderboard maintainer scored the entry on his own harness; we self-reported
nothing. 
Our entry, internal run \sysname{vh5\_0d}, placed first on three axes. It reached 9.02\% word
error rate (WER) on MediaSpeech, against 11.93\% for the best unmodified model,
\sysname{whisper-large-v3}. 
It reached 8.73\% on the rebuilt Common Voice~17 test, against 9.53\%. 
And it reached 9.71\% on the five-set average, against 9.90\%. 
These are the figures of the 2026-06-27 store update, re-verified on 2026-07-04. 

The board also keeps a private held-out set, Yeni Split~0 (YS-0). It runs to roughly 18 hours
that its maintainer describes as ``noisy real-world recordings''. 
There the ranking inverted. The unmodified \sysname{whisper-large-v3} stayed first at 11.35\%,
and our entry fell to fourth at 12.82\%. 
On the axis that most resembles deployment, we were beaten by the untouched base model, and by no rival fine-tune.

We name this failure mode \emph{acoustic narrowing} and define it here. It is a
fine-tuning-induced contraction of the acoustic distribution a model serves well: in-domain WER
improves, while WER on unseen distributions falls below that of the unmodified base model.
The unit that contracts is an acoustic-domain bundle as it occurs in practice, not an isolated
acoustic variable; Section~\ref{sec:crosslingual} states what our design does and does not
separate.
Every ingredient has published precedent. Fine-tuning distorts pretrained features inside the
span of the adaptation data~\citep{kumar2022finetuning}. Zero-shot Whisper is robust precisely
because it adapts to no single benchmark~\citep{radford2023whisper}. The loss has
been measured in speech for Irish and European Portuguese~\citep{qian2024learn,carvalho2026merging}.
Section~\ref{sec:related} sets out what each of those threads establishes and where the present
result sits against it.

A single private test set could be an accident of composition; a second, independently built one is harder to dismiss.
On a benchmark constructed by another team, the National Software Certification Center (UYSM), covering 100 FLEURS-TR~\citep{conneau2023fleurs} utterances under 20 acoustic conditions (clean, additive noise, Opus compression, and their combinations), 
the unmodified turbo beat every one of our fine-tunes on every condition group. 
Two test distributions, one private and one public, expose the same phenomenon.
The choice of Turkish is also material: most sizable Turkish corpora remain closed, so public
evaluation concentrates on a few clean read- and broadcast-speech sets
(Section~\ref{sec:related}).

Our contributions are fivefold.

\textbf{(1) A documented case of leaderboard-driven acoustic narrowing}, established on public
data. The fine-tune wins exactly the boards its corpus covered and loses the rest
(Figure~\ref{fig:board}). On another team's twenty-condition FLEURS-TR matrix, the unmodified
base beats every fine-tune in every condition group (Figure~\ref{fig:fleurs}). 

\textbf{(2) Proxy blindness}, the most portable result. The instrument a practitioner would
reach for to \emph{prevent} narrowing conceals it instead. A selection proxy built from the
pools that feed training inherits their coverage. It therefore ranks the narrowed fine-tune
above the more robust base, inverting the external verdict (Table~\ref{tab:proxy},
Figure~\ref{fig:proxy}). Accuracy on the reference distribution cannot reveal shift-induced
damage~\citep{wortsman2022wiseft}, so the corrective is a proxy held out from an
\emph{independently produced} distribution. 

\textbf{(3) A recovery recipe}, fixed in advance and executed in full: an 875.2-hour
mixture with per-source provenance disclosed, a higher-capacity base, and weight-space robustness anchored to the
raw base~\citep{wortsman2022wiseft,wortsman2022modelsoups}. The trajectory-averaged fine-tune
holds a decisive margin on three same-provenance held-out axes. It also returns to parity on the
FLEURS-TR test split, 8.16\% against 8.26\% with a confidence interval of $[-0.20,+0.40]$, where
every predecessor lost by whole points. That split is held out from training, but it is drawn
from the distribution whose development split drove early stopping. Within the run, trajectory
averaging is the strongest candidate under a fixed budget
(Table~\ref{tab:recovery}). 

\textbf{(4) Two experiments that bound the claim.} These belong in the same paper as the
recovery, because without them a claim that some pipeline ``arrests narrowing'' cannot be
falsified: one needs the conditions under which narrowing does not occur at all. Replicating the design in Kazakh reproduces
the inversion on corpora we did not build, with damage ordered by acoustic distance from the
pool. A concurrent Kazakh study, published before we looked, shows the same inversion in its own
numbers without naming it~\citep{yeshpanov2026songs}. 

A controlled architecture test then \emph{locates} the effect. Pool and probe come from one
corpus absent from both models' pretraining, and adaptation capacity is matched. Under those
conditions the attention encoder--decoder loses 2.23 points out of pool, while a CTC
encoder-only model \emph{gains} 3.46, each replicated across three seeds. Narrowing is therefore
not an inevitable consequence of fine-tuning: under this protocol it depends on the system being
adapted. The decoder is our leading explanation, though we did not isolate it. An RNN-T arm,
autoregressive without attention, is the experiment that would settle
it. 

\textbf{(5) TurkSpeak Bench}, a robustness-first Turkish ASR leaderboard scoring models under
the same twenty-condition grid used in our validation. 

\phead{Scope beyond speech}
Nothing in the proxy-blindness argument depends on audio: a proxy drawn from the training pools
cannot reveal what those pools omit, whatever the modality. We state this as a conjecture, since
we tested it only in speech.


\section{Related work}
\label{sec:related}

\phead{Benchmark-directed optimization}
That leaderboard standing and held-out reality diverge is established outside speech and
inside it. Image classifiers that dominate ImageNet lose substantial accuracy on a faithfully re-collected
test set~\citep{recht2019imagenet}. The ESB benchmark was designed on the premise that
dataset-specific tuning presumes knowledge of the test distribution~\citep{gandhi2022esb}.
Transcripts of standard ASR evaluation sets circulate verbatim in language-model pretraining
corpora~\citep{tseng2025contamination}. Published Swiss German ASR results dissolve once
contamination and transcription convention are controlled~\citep{akeret2026swissgerman}. This literature establishes that benchmark scores drift from deployment behaviour. It stops short of attributing a specific ranking inversion to a specific adaptation decision, with the \emph{unmodified base model} as the
competitor that wins, under scoring performed by a third party. That is the case we document.

\phead{Fine-tuning distortion and forgetting}
The mechanism has a general statement and speech-specific instances. Fine-tuning distorts
pretrained features inside the span of the adaptation data, an out-of-distribution penalty that
early stopping cannot repair~\citep{kumar2022finetuning}; zero-shot Whisper approaches
human-level effective robustness precisely because it adapts to no single
benchmark~\citep{radford2023whisper}. In speech, fine-tuning \sysname{whisper-large-v3} on
twelve hours of Irish raises its LibriSpeech WER from 3.7\% to 8.8\%~\citep{qian2024learn}, and
joint fine-tuning on European Portuguese corpora degrades Brazilian Portuguese, English and
multilingual FLEURS below the zero-shot model~\citep{carvalho2026merging}. These papers measure
the loss. None of them reports that the instrument a practitioner would use to \emph{detect}
the loss inverts the verdict. That is the result we consider most portable
(Section~\ref{sec:results}).

\phead{Weight-space robustness}
Our recovery stage uses published tools: linear interpolation between a
fine-tune and its base~\citep{wortsman2022wiseft}, averaging of independently fine-tuned
weights~\citep{wortsman2022modelsoups}, checkpoint averaging along a training
trajectory~\citep{izmailov2018averaging,karita2019comparative}, and task-vector
editing~\citep{ilharco2023editing}. The closest speech result is
\citet{ducorroy2025merging}, where uniform checkpoint averaging improves fine-tuned Whisper on
disordered speech at no inference cost, and where the gain is capacity-dependent in a way that
motivates our choice of base. Our contribution here is evaluation-side: which of these candidates survives a probe of independent provenance
under an acceptance rule fixed before the candidates existed
(Section~\ref{sec:recovery-weights}).

\phead{Low-resource ASR evaluation}
Turkish sharpens the problem for a structural reason. The Open ASR Leaderboard's multilingual track covers a
handful of European languages and no Turkish~\citep{srivastav2025openasr}; most sizable Turkish
corpora remain closed~\citep{mussakhojayeva2023multilingual}, which concentrates public
evaluation, and therefore fine-tuning, on a few clean read- and broadcast-speech sets. The
effect we name has already been recorded in miniature in this literature: models fine-tuned only
on Common Voice fail on unseen call-centre audio~\citep{mercan2023performance}, and a Whisper
LoRA fine-tune improved Common Voice while degrading an unseen commercial test
set~\citep{polat2024whisper}. A concurrent Kazakh study reports the same inversion in its own
numbers without naming it~\citep{yeshpanov2026songs}. The literature has lacked a leaderboard-scale account scored by a third party, a name for the failure mode, and a tested route out; Section~\ref{sec:crosslingual} shows the account is not language-specific.

\section{Anatomy of a Leaderboard Win}
\label{sec:diagnosis}

We now dissect the win: what the board measures, what our entry did to top it, and how the
board's own private half first raised the alarm. A framing point governs the whole section. The private split is scored on the maintainer's harness. That is a strength,
but its contents are not published and we cannot re-run models on it, a decisive
limitation. We therefore read this section as the observation that prompted the investigation.
The claim that fine-tuning narrowed the model is settled in Section~\ref{sec:validation} on
public data under controlled conditions, and again in Section~\ref{sec:crosslingual} in a
second language and across two architectures.

The ysdede Turkish ASR leaderboard is a community-run public benchmark on
Hugging Face~\citep{ysdede2026leaderboard}. It ranks models on five principal evaluation sets (three public, two private held-out) plus two aggregate views over them (a sixth, sparsely populated low-quality private split lists only two maintainer models and none of the models compared here). These seven boards (Table~\ref{tab:board}) carry all the evidence in this section. 
MediaSpeech supplies 10~h of Turkish speech collected from official media YouTube channels and manually transcribed~\citep{kolobov2021mediaspeech}. 
Common Voice~17 \emph{fixed} (cv17-fixed) is a CC0 rebuild of Mozilla Common
Voice~17 Turkish~\citep{ardila2020commonvoice}, resegmented with the
maintainer's ASRTK pipeline (Silero VAD, 16~kHz, sub-second clips dropped)~\citep{asrtk2024}. Its test split holds 9{,}650 clips (7.33~h). 
The third public set, TurkishVoice, is synthetic: news audio generated with
Microsoft text-to-speech~\citep{turkishvoice2024}: one of the board's five sets contains no real speech, a fact that sharpens the coverage analysis
below. 
The private pair consists of tr-med-audio, 7.97~h of medical-domain speech, 
and Yeni Split~0 (YS-0), 17.95~h across 5{,}936 samples that the maintainer's public documentation describes as ``noisy real-world recordings''. 
The board is completed by two aggregates. \emph{Unseen} is the mean over the two private sets, listing a model only if it has both; 
\emph{Average} is the mean over whatever rows a model happens to have, so its composition varies, a caveat Table~\ref{tab:board} makes explicit. 

Scoring is uniform and, importantly for what follows, independent. Every row
was produced by the maintainer's own harness: WhisperX over faster-whisper/CTranslate2, float16, conditioning on previous text disabled,
the Turkish-specific trnorm normalizer, on an NVIDIA L4 GPU, and WER ranks the boards. 
We submitted weights. The maintainer ran them. 
All board numbers in this paper are as of the 2026-06-27 store update, re-verified live on 2026-07-04. 

Our entry (public row \repo{RsGoksel\_ITU\_Mainframe}, internal run
\sysname{vh5\_0d}) is a LoRA fine-tune~\citep{hu2022lora} of
\sysname{whisper-large-v3-turbo} with rank $r{=}32$, $\alpha{=}64$ on the query and value projections, trained on 997.2~h of Turkish audio (688{,}397 clips). 
The mixture is dominated by a single community podcast aggregate at 53.7\% (535.8~h). The largest remaining components are human-transcribed YouTube speech from the Turkish Speech Corpus (209.3~h), Khan Academy Turkish lectures (104.2~h and 52.1~h), and 56.8~h drawn from cv17-fixed, the training side of a board set. 
The corpus contains no TurkishVoice audio and, the sets being private, YS-0 and tr-med-audio could not have been deliberately included (incidental overlap cannot be audited, see Section~\ref{sec:limitations}). 

\begin{figure}[t]
\sidecapoff
\centering
\includegraphics[width=0.82\textwidth]{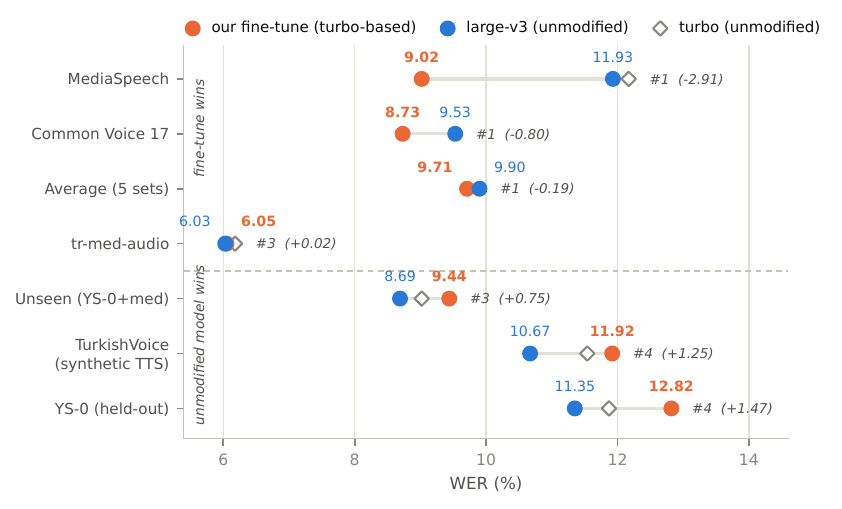}
\caption{The leaderboard inversion. WER of our entry \sysname{vh5\_0d} (orange)
and unmodified \sysname{whisper-large-v3} (blue) on the seven boards of the
ysdede Turkish ASR leaderboard (Table~\ref{tab:board}), sorted by $\Delta=$ ours $-$ large-v3; white
diamonds mark unmodified turbo where the store holds a composition-matched
row. Annotations give our rank and $\Delta$. The fine-tune wins every board its 997.2~h corpus covered, loses the private noisy split and the synthetic-TTS board, and only reaches parity on the private medical set. 
}
\label{fig:board}
\end{figure}

The outcome, summarized in Table~\ref{tab:board} and
Figure~\ref{fig:board}, splits cleanly in two. On the covered boards we win outright. We take \#1 on MediaSpeech at 9.02\% WER, a 2.91-point
margin over the best unmodified checkpoint, \sysname{whisper-large-v3} at 11.93\%. 
We take \#1 on cv17-fixed at 8.73\% against 9.53\%. That margin should be read with care: the
rebuild has documented speaker leakage across Common Voice splits, and 56.8~h of its train side
sits in our corpus. 
We also take \#1 on the five-set Average at 9.71\% against 9.90\%. 
By the board's headline views, ours was the strongest Turkish ASR model
listed. On the boards the corpus did not cover, the ranking inverts. On YS-0 we fall to \#4 at 12.82\%. Three unmodified checkpoints finish ahead of us: large-v3 at
11.35\%, turbo at 11.87\% and large-v2 at 12.30\%. The second of those is our own base. 
On the Unseen aggregate we are \#3 at 9.44\% against large-v3's 8.69\% and turbo's 9.02\%; 
on synthetic TurkishVoice, \#4 at 11.92\% vs.\ large-v3's 10.67\%. 
Fine-tuning bought 2.91 points on MediaSpeech over the best unmodified checkpoint (3.15 over its own base, turbo) at a cost of 0.95 points on YS-0 relative to that base: in-domain improvement paired with regression below the unmodified base on an uncovered distribution. This is the shape the next section establishes on data we can re-score. 

The win--loss pattern tracks coverage. Ordering the seven boards by $\Delta$ (Table~\ref{tab:board}) gives $-2.91$ MediaSpeech, $-0.80$ cv17-fixed, $-0.19$ Average, $+0.02$ tr-med-audio, $+0.75$ Unseen, $+1.25$ TurkishVoice, $+1.47$ YS-0. 
The two boards we win decisively are exactly the two whose distributions the training corpus covered: media and podcast speech, and cv17-fixed through its own train split. The two where we fall to \#4 are exactly the two it did not cover: synthetic TTS prosody and the private noisy split. 
One apparent exception is near-parity on tr-med-audio, 6.05\% against 6.03\%, with zero hours of
medical training audio. It does not support a generalization reading. Large-v3, our fine-tune and
turbo sit within 0.15 points of one another there, at 6.03\%, 6.05\% and 6.18\%. That pattern fits
an easy set near saturation for large models. Transfer would look different. 
The fine-tune wins where its training distribution covered the test
distribution, and loses everywhere else.

The board carries a second signal, orthogonal to coverage: capacity. Restricted to unmodified checkpoints on YS-0, large-v3 beats turbo at 11.35\% vs.\ 11.87\%, 
and the two differ chiefly in decoder depth, 32 layers against 4~\citep{radford2023whisper,openai2024turbo}. 
\citet{ducorroy2025merging} report a compatible asymmetry in a different regime: uniform weight merging recovers substantial WER on fine-tuned large-v3 but almost none on turbo (20.8\% to 20.3\%), on dysarthric-English development data. 
Both observations point the same way, and Section~\ref{sec:recovery} acts on
them: we rebase recovery on large-v3.

Finally, the episode explains why we did not catch the narrowing ourselves. Model selection for the planned re-submission leaned on internal 120-clip proxies (2026-06-29), including a ws-proxy for the YS-0 axis assembled from our in-house broadcast pool. 
That proxy ranked \sysname{md\_v1} ahead of both unmodified checkpoints, 7.52\% against 7.79\%
for raw large-v3 and 8.00\% for raw turbo. 
\sysname{md\_v1} is the diversified successor fine-tune of Table~\ref{tab:lineage}; we pushed its
weights to the submission repository that same day, and the board never re-scored
them. 
The board's actual YS-0 result placed raw large-v3 1.47 points ahead of our entry. 
A proxy built from the same pools that feed training inherits the training
distribution's coverage biases and systematically understates
narrowing. This is the ASR analogue of the observation, made for image
classifiers, that fine-tunes with near-identical reference-distribution
accuracy can differ dramatically under distribution shift, so in-distribution
scores alone cannot reveal the damage~\citep{wortsman2022wiseft}. The diagnosis forces two design changes. Selection must add a proxy held out from a
human-transcribed corpus outside the board's own sets (Section~\ref{sec:recovery}) and, more
importantly, an evaluation on an independently produced distribution; a
public robustness-first benchmark, TurkSpeak~Bench (Section~\ref{sec:bench}).

\section{Independent Cross-Team Validation}
\label{sec:validation}

A diagnosis that rests on a single private test set is open to two objections: the set itself may
be idiosyncratic, and the team that trained the models also picked the measurement. We therefore
sought replication on a benchmark we did not build: another team, the National Software
Certification Center (\emph{Ulusal Yazılım Sertifikasyon Merkezi}, UYSM), constructed one 
from a public corpus the leaderboard does not score and supplied human-verified references. We
scored the checkpoints we supplied under its protocol, the partner ran its own reference models, and
the two teams reconciled text normalizers (below).
UYSM is a research and certification centre at the author's institution and provided the
computing hardware used in
this work. It had no part in choosing our models, mixtures or checkpoints, and we had none in
designing its benchmark, selecting its conditions or scoring its reference models. Where we call
this evidence cross-team, that division of labour is what we mean by the term. The probe is FLEURS-TR, the Turkish portion of the 102-language FLEURS
benchmark~\citep{conneau2023fleurs}. It consists of prompted read speech recorded from
human-translated Wikipedia sentences, and is released with a speaker-disjoint test
split~\citep{conneau2023fleurs}. 
Acoustically and stylistically it sits at a distance from the board's broadcast and
synthetic-TTS material. It is also longer-form than the board's Common Voice prompts, at a mean
of 12.2\,s, and far from the podcast-heavy corpus behind our board entry. 
If the inversion of Figure~\ref{fig:board} were an artifact of YS-0 or of the board's
harness, it would vanish here. In contrast, it survives.

The team drew 100 utterances from the FLEURS-TR test split, ranging from 4.2 to 24.4\,s with a
mean of 12.2\,s. Each was rendered under 20 acoustic conditions, giving roughly 20 minutes of
audio per condition. The conditions comprise the clean originals,
MUSAN~\citep{snyder2015musan} noise at 20, 10 and 0\,dB SNR, Opus transcoding at 32, 16, 12 and
8\,kbps, and the twelve noise$\times$codec combinations. 
Every reference transcript was human-verified, with 0 of 100 clips flagged. 
Scoring mirrors the public board's stack~\citep{ysdede2026leaderboard}. It reports corpus-level
WER, applies the \code{trnorm} Turkish text normalizer to references and hypotheses alike,
and decodes with faster-whisper (CTranslate2) at beam size 5. That is the same backend family
the board uses, chosen precisely so that implementation differences cannot confound the
comparison. 
The evaluation was completed on 2026-07-02. The same degradation grid later powers TurkSpeak~Bench. 

\begin{figure}[t]
\sidecapoff
\centering
\includegraphics[width=0.86\linewidth]{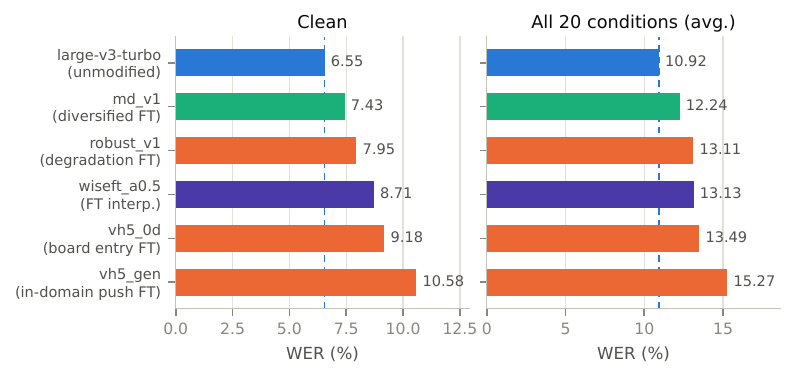}
\caption{Cross-team validation on FLEURS-TR: clean (left) and all-20-condition 
average (right) corpus-WER for the unmodified \sysname{large-v3-turbo} and the five fine-tuned
variants in its lineage, on the degradation benchmark built by another team (UYSM). The 
dashed line marks the
unmodified base; every fine-tune falls short of it in both panels, and the deficit grows with the
narrowness of the fine-tuning signal. Full matrix, including the \sysname{whisper-base} reference,
in Table~\ref{tab:fleurs}.}
\label{fig:fleurs}
\end{figure}

\begin{table}[t]
\centering
\caption{Cross-team FLEURS-TR validation (UYSM): corpus-WER (\%) on 100 test utterances under 20
acoustic conditions, grouped as the clean originals; MUSAN noise at $\{20,10,0\}$\,dB SNR; Opus
at $\{32,16,12,8\}$\,kbps; their 12 combinations; and all 20. Human-verified references;
\code{trnorm} on references and hypotheses; faster-whisper (CTranslate2), beam~5.
\sysname{whisper-base} is the partner team's reference model. Best per column in bold.
The last column gives each fine-tune's ALL-20 deficit against the unmodified base with a 95\%
paired bootstrap interval ($B{=}10{,}000$), resampled by utterance rather than by
utterance-condition cell. The text below states the re-decode these
intervals come from and the two repetition-loop decodes excluded from it. 
}
\label{tab:fleurs}
\small
\setlength{\tabcolsep}{4.4pt}
\begin{tabular}{@{}lrrrrrc@{}}
\toprule
& & & & & & $\Delta$ ALL-20 vs.\ base \\
Model & Clean & Noise & Codec & Noise{+}Codec & ALL-20 & [95\% CI] \\
\midrule
\sysname{large-v3-turbo} (unmodified)                & \textbf{6.55} & \textbf{9.12} & \textbf{7.11} & \textbf{13.01} & \textbf{10.92} & --- \\ 
\sysname{md\_v1} (diversified FT)                    & 7.43  & 10.66 & 8.30  & 14.35 & 12.24 & $+1.32$ $[+0.19,+2.69]$ \\ 
\sysname{robust\_v1} (degradation FT)                & 7.95  & 11.91 & 8.95  & 15.22 & 13.11 & $+2.18$ $[+0.79,+3.93]$ \\ 
\sysname{wiseft\_a0.5} (FT$\leftrightarrow$FT interp.) & 8.71  & 11.52 & 9.17  & 15.22 & 13.13 & $+2.20$ $[+0.79,+3.97]$ \\ 
\sysname{vh5\_0d} (board entry)                      & 9.18  & 12.03 & 9.36  & 15.59 & 13.49 & $+2.56$ $[+1.14,+4.22]$ \\ 
\sysname{vh5\_gen} (in-domain push)                  & 10.58 & 14.33 & 11.04 & 17.31 & 15.27 & $+4.35$ $[+2.87,+6.16]$ \\ 
\midrule
\sysname{whisper-base} (partner reference)           & 26.14 & 38.81 & 30.22 & 48.18 & 42.08 & --- \\ 
\bottomrule
\end{tabular}
\end{table}

Table~\ref{tab:fleurs} reports the full matrix. Figure~\ref{fig:fleurs} shows the clean and ALL-20
views. The unmodified turbo beats every one of our fine-tunes in every condition group: 
6.55\% WER clean and 10.92\% across all twenty conditions, against 9.18\% and 13.49\% for 
\sysname{vh5\_0d}, the entry that tops the public MediaSpeech board. The partner's 
\sysname{whisper-base} reference row (26.14\% clean, 42.08\% ALL-20) anchors the scale and confirms 
the pipeline stresses models in the expected order.
Those intervals come from a re-decode of the grid under this protocol, whose deficits agree with
the published columns to within 0.05 points on every row, so each interval and the deficit it
brackets are measured on the same decode. Two of its 12{,}000 decodes entered a repetition loop,
flagged by the uniform rule that insertions exceed twice the reference length, and are excluded;
including them widens two intervals to $[+1.20,+5.16]$ and $[+2.99,+6.53]$ and changes no
sign. 
The deficit is not a reading of point estimates alone. Resampling the hundred utterances, all
five ALL-20 intervals exclude zero, from $[+0.19,+2.69]$ for the mildest fine-tune to
$[+2.87,+6.16]$ for the narrowest. The ordering is therefore established, not merely
observed. 
On the clean condition taken by itself the two mildest fine-tunes fall short of the base without
clearing zero individually ($[-0.54,+2.42]$ and $[-0.18,+3.09]$), which is what one hundred
utterances of a single condition can support; the twenty-condition aggregate is where this
table's evidence sits. 
This is the zero-shot-versus-fine-tuned robustness gap of \citet{radford2023whisper}, reproduced within a single model family: FLEURS is
the very suite on which Whisper's zero-shot multilingual coverage was 
charted~\citep{radford2023whisper}, so read speech of this kind lies well inside the raw model's weakly supervised distribution, and, after our fine-tuning, outside it. FLEURS was not even
wholly unseen: \sysname{vh5\_0d}'s corpus included 8.3\,h of FLEURS speech within its 
997.2\,h, and its early stopping monitored the FLEURS dev split. A sliver of coverage and a 
dev-set gate did not prevent the contraction.

The ordering among the fine-tunes tracks the narrowness of their training signal. Worst is
\sysname{vh5\_gen} at 10.58\% clean and 15.27\% ALL-20. It was fine-tuned from \sysname{vh5\_0d} on 
a mixture dominated by the board's synthetic-TTS news domain and early-stopped on a development
set of 2{,}000 clips drawn from that same domain, so optimization and model selection both pointed 
inward. Closest to the base is \sysname{md\_v1}, the most diversified mixture in the lineage, at 7.43\% clean and 12.24\% ALL-20: the most diversified mixture shows roughly half of 
\sysname{vh5\_0d}'s 2.56-point ALL-20 deficit, but only half. \sysname{robust\_v1} and 
\sysname{wiseft\_a0.5} land in between (13.11 and 13.13 ALL-20). That interpolating two fine-tunes 
buys essentially nothing here (8.71\% clean) previews Section~\ref{sec:recovery}. Narrowing is 
graded: the ordering runs with in-domain pressure, though these checkpoints differ in more than their mixtures.

Cross-laboratory WER comparisons can founder on text normalization, so the two teams reconciled
normalizers explicitly. Re-scoring \sysname{md\_v1} with the partner's independent normalizer gives
7.77\% clean and 12.61\% ALL-20 against 7.43\% and 12.24\% under ours, and the 
\sysname{whisper-base} row reconciles to within 0.06 points (26.08 vs.\ 26.14). Normalizer choice 
is thus a ${\sim}0.3$-point effect: far too small to produce \sysname{vh5\_0d}'s 2.56-point ALL-20 
gap to its base, and smaller than every fine-tune's deficit in every condition group. The same 
result emerges from two teams running two harnesses on two test distributions. The
leaderboard's private YS-0, described by its
maintainer as noisy real-world recordings, and the public read-speech FLEURS-TR, expose the same 
phenomenon. The fine-tune that wins the public boards is the wrong model to deploy on audio those
boards never showed it.

\section{Recovery: Diversify, Upsize, Average}
\label{sec:recovery}

The diagnosis is specific enough to act on, and it points at three distinct faults. Our board
entry won exactly the boards covered by its 997.2\,h corpus, 53.7\% of which was podcast audio,
and lost the rest. 
Its base was the reduced-decoder \sysname{turbo}, which already trails \sysname{large-v3} on the
held-out split \emph{before} any fine-tuning. 
Finally, every selection decision after the board entry was referenced to in-domain data rather
than to the base model. The entry itself early-stopped on a FLEURS-dev signal whose corpus its
own training data contaminated, to the extent of 8.3\,h of
FLEURS. 
We move three levers together. The design is a bundle, and we say so. No lever here is separately identified: rebuild the training mixture to cover the unseen
acoustic support (\S\ref{sec:recovery-data}), fine-tune from the higher-capacity base
(\S\ref{sec:recovery-capacity}), and impose robustness in weight space anchored to the raw
base model (\S\ref{sec:recovery-weights}). The methodology below was fixed before launch. The
mixture build and the large-v3 run were then executed on 2026-07-07, 
and Table~\ref{tab:recovery} reports the measured ablation.

\subsection{Data Diversification}
\label{sec:recovery-data}

Acoustic narrowing is, on our evidence, a coverage failure, so the first fix is a change in which hours, and not how many: extend the mixture's support toward the deployment-relevant unseen distribution and away from the already-covered boards.

The target is stated without reference to any private set, and deliberately so: a recipe
justified by guessing at an unauditable distribution could not be evaluated by anyone else.
The measured evidence says that our fine-tunes lose wherever speech is spontaneous or degraded, in every condition group of the FLEURS-TR grid, and the deficit persists across clean, noise, codec and combined groups (Table~\ref{tab:fleurs}), while the corpus that produced them was 53.7\% a single
podcast aggregate. 
The fix follows from that alone: extend coverage toward spontaneous, multi-speaker,
variably-recorded speech with verifiable transcription provenance. For completeness we note
that the board owner describes the private split as roughly 18\,h of noisy real-world
recordings and that their public tooling suggests a spontaneous-YouTube,
VAD-segmented profile~\citep{asrtk2024}. This suggests that the direction we chose is sound, although it is
inference about a set we cannot inspect, and nothing in the recipe rests on
it. 

We therefore rebuilt the mixture around spontaneous YouTube speech with verified transcription
provenance (Table~\ref{tab:data}, Figure~\ref{fig:mix}). The ISSAI Turkish Speech Corpus (TSC)
contributes 218.2\,h / 186{,}171 utterances of news, interviews, talk shows, and 
documentaries, transcribed by a team of language specialists: genuinely human-transcribed
YouTube speech~\citep{mussakhojayeva2023multilingual}. 
YODAS \sysname{tr000} contributes 588.7\,h of user-uploaded caption data~\citep{li2023yodas}; 
we exclude the 4{,}067.7\,h auto-caption shard \sysname{tr100}: in the corpus paper's English 
comparison, manual-caption training reaches 14.9 CER where auto-caption training collapses to 
32.3, deletions dominating. 
We also repeat the card's caveat. A ``manual'' caption is user-uploaded, \emph{not}
necessarily transcribed by a human. 
Common Voice 22 adds CC0 crowd-read speech, 128.01 validated hours, of which our on-disk 
subset is a partial 55.2\,h download~\citep{ardila2020commonvoice,commonvoice22}. 
A uniform quality filter (1--30\,s duration, $\geq$3 characters/s) yields 572.9\,h of YODAS 
and 213.2\,h of TSC; 
from the latter we carve a 5.0\,h / 4{,}181-clip held-out slice, \sysname{tsc\_heldout}, as an 
unseen-profile selection proxy, and cap the TSC train share at 200\,h. 

Contamination discipline precedes scale. Four rules govern the build. We maintain dual
text-normalized exclusion sets over TurkishVoice and CV17-test references. We remove FLEURS from
training entirely, retaining the dev split only for early stopping and holding the test split as
an out-of-distribution probe~\citep{conneau2023fleurs}. We deduplicate TSC audio by basename.
And we run a hard overlap gate before any launch. 
In the dry-run build these rules drop $2{,}234 + 207 + 13{,}351$ clips. 

The dry run totals 410{,}903 clips and 673.6\,h. 
The final build, executed on 2026-07-07, raises YODAS to 197.4\,h, its 200\,h cap after the
quality filter. It raises the TSC train share to 184.0\,h, where the held-out carve-out and
deduplication keep it under the same cap, and CV22 to 53.5\,h. The manifest as built measures
\textbf{592{,}836 clips and 875.2\,h} (Figure~\ref{fig:mix}). 
The 896\,h projection from cap arithmetic overshot by $\sim$21\,h, because caps interact with
carve-outs and filters. 
One disclosure survives diversification. The largest single component is still 253.8\,h of
synthetic Microsoft-TTS news, the board's own in-domain set and 29.0\% of the final build
and carries a non-commercial license. A further 17.2\,h is synthetic medical
TTS. 
Figure~\ref{fig:mix} flags both. The trajectory across mixes nonetheless moves mass toward
licensed, human- or user-transcribed spontaneous speech.

\begin{table}[t]
  \centering
  \footnotesize
  \setlength{\tabcolsep}{3.5pt}
  \caption{Data sources of the recovery mixture. Hours trace corpus $\rightarrow$ post-filter
  (1--30\,s, $\geq$3 chars/s) $\rightarrow$ \sysname{mix-v2} dry-run build; 
  the final build (2026-07-07) raises YODAS to 197.4\,h, the TSC train share to 184.0\,h, and 
  CV22 to 53.5\,h, measuring 592{,}836 clips / 875.2\,h in total. 
  A 5.0\,h TSC slice is held out as the \sysname{tsc\_heldout} selection proxy. 
  ``n/s'' = not stated in the primary source; ``unverified'' entries are carried to the
  Limitations.}
  \label{tab:data}
  \setlength{\tabcolsep}{3pt}%
  \begin{tabular}{@{}L{0.212\linewidth}L{0.162\linewidth}rrrL{0.168\linewidth}L{0.098\linewidth}@{}}
    \toprule
    & & \multicolumn{3}{c}{Hours} & & \\
    \cmidrule(lr){3-5}
    Source & Domain / type & Corpus & Filt. & Mix (dry run) & Transcription provenance & License \\
    \midrule
    TurkishVoice (rest)~\citep{turkishvoice2024} & synthetic TTS news & n/s & n/a & 253.8 & synthetic (Microsoft TTS) & CC-BY-NC-4.0 \\ 
    ISSAI TSC~\citep{mussakhojayeva2023multilingual} & YouTube broadcast (news, interviews, talk shows, documentaries) & 218.2 & 213.2 & 184.1 & human (specialist team) & MIT\,/\,CC-BY-4.0 \\ 
    worldspeech (in-house) & parliament / broadcast aggregation & 610 & n/a & 60.0 & n/s & in-house \\ 
    merge (community aggregate) & podcast aggregate & $\sim$300--500 & n/a & 42.2 & unverified & unverified \\ 
    YODAS \sysname{tr000}~\citep{li2023yodas} & spontaneous YouTube, multi-domain & 588.7 & 572.9 & 36.1 & user-uploaded captions (not nec.\ human) & CC-BY-3.0 \\ 
    Degradation-augmented & telephone / noise / reverb copies & n/a & n/a & 32.2 & inherited from sources & n/a \\ 
    medv3 (synthetic) & medical TTS & n/s & n/a & 17.2 & synthetic TTS & n/s \\ 
    issai (earlier TSC subset) & as TSC above & n/a & n/a & 16.9 & human (specialist team) & MIT \\ 
    Common Voice 22~\citep{ardila2020commonvoice,commonvoice22} & crowd-read sentences & 128.01 & 55.2 & 13.1 & prompted read, community-validated & CC0 \\ 
    Khan Academy TR & educational lectures & n/s & n/a & 12.3 & n/s & n/s \\ 
    CV17-fixed & read sentences (CV17 rebuild) & n/s & n/a & 3.4 & prompted read, community-validated & CC0 \\ 
    others & misc. & n/a & n/a & 2.4 & n/a & n/a \\ 
    \midrule
    \multicolumn{4}{@{}l}{Total, dry-run build} & 673.6 & & \\ 
    \multicolumn{4}{@{}l}{Final build, measured (YODAS 197.4, TSC 184.0, CV22 53.5)} & 875.2 & & \\ 
    \bottomrule
  \end{tabular}
\end{table}

\subsection{Capacity}
\label{sec:recovery-capacity}

The second lever is the base model, and we justify it from evidence a reader can check.
Published ASR work reports that weight-space merging gains concentrate in larger
models~\citep{ducorroy2025merging}, and on the FLEURS-TR grid capacity is not a neutral choice:
the smaller \sysname{whisper-base} trails turbo by fifteen to thirty points across condition
groups (Table~\ref{tab:fleurs}). 
That comparison does not reach large-v3. The board's private split does, since raw
\sysname{large-v3} outranks raw \sysname{turbo} there, but that is corroboration only. 
\sysname{vh5\_0d} fine-tuned
\sysname{whisper-large-v3-turbo}, $\sim$809M parameters with a 4-layer decoder. The recovery 
run fine-tunes \sysname{whisper-large-v3}, $\sim$1.55B parameters with 32 decoder 
layers~\citep{radford2023whisper,openai2024turbo}. 
The added compute is justified by two observations. Before any fine-tuning, raw
\sysname{large-v3} already beats raw \sysname{turbo} on YS-0, 11.35\% vs.\ 11.87\% WER 
(Table~\ref{tab:board}): decoder capacity correlates with retention on the held-out
distribution. 
The weight-averaging methods of \S\ref{sec:recovery-weights} also appear
capacity-dependent: \citet{ducorroy2025merging} report that merging cuts WER from 15.0 to 13.2 on 
\sysname{large-v3} yet barely moves \sysname{turbo}, 20.8 to 20.3, numbers obtained on 
dysarthric-speech dev subsets, not on acoustic-OOD benchmarks, but the asymmetry is the
point. 
Both read naturally through the plasticity--stability trade-off documented for Whisper
adaptation~\citep{qian2024learn}; on that reading, the smaller the decoder, the less room for
specialization to coexist with broad competence. The full recipe and lineage appear in
Table~\ref{tab:lineage}.

\subsection{Weight-Space Robustness}
\label{sec:recovery-weights}

Fine-tuning is a single direction in weight space,
$\tau = \theta_\mathrm{FT} - \theta_0$~\citep{ilharco2023editing}, and applying it at partial
strength interpolates between base and fine-tune. WiSE-FT~\citep{wortsman2022wiseft} makes this
the robustness mechanism:
\begin{equation}
  \theta(\alpha) \;=\; \alpha\,\theta_\mathrm{FT} + (1-\alpha)\,\theta_0,
  \qquad \alpha \in [0,1].
  \label{eq:wiseft}
\end{equation}
In vision this recovers most zero-shot robustness while maintaining target accuracy, at no extra
training or inference cost. It is well posed because same-initialization endpoints are connected
by a linear high-accuracy path~\citep{wortsman2022wiseft}.
Our sweep interpolates the large-v3 fine-tune toward the \emph{raw} base with
$\alpha \in \{0.3, 0.4, 0.5, 0.6, 0.7\}$, executed alongside the fine-tune. 
It turned out to be insurance we did not need. WER degrades monotonically toward the base on the
primary \sysname{tsc\_heldout} proxy, and the pure fine-tune is the champion
(Table~\ref{tab:recovery}). 

The qualifier ``toward the raw base'' is load-bearing, and we learned it the hard way:
our earlier \sysname{wiseft\_a0.5} averaged two fine-tunes, $0.5\cdot$ 
\sysname{robust\_v1} (the degradation-robust fine-tune of Table~\ref{tab:lineage}) $+\;0.5\cdot$ 
\sysname{vh5\_0d}, and never re-approached $\theta_0$. 
On FLEURS-TR it merely landed between its endpoints, at 8.71\% clean WER against 7.95\% and
9.18\%. 
Over all 20 conditions it trailed \sysname{md\_v1}, 13.13\% against 12.24\%
(Table~\ref{tab:fleurs}). 
There is no synergy here: both endpoints share the narrowed lineage, so their average stays
inside the narrowed region (Figure~\ref{fig:weightspace}).

Two further averaging steps complement interpolation. Both were executed on 2026-07-18 and
measured in Table~\ref{tab:recovery}. 
Within a run, we average the saved checkpoints; trajectory averaging is standard end-to-end
ASR practice: ESPnet Transformer recipes ship a
last-10-epoch average in place of early stopping~\citep{karita2019comparative} 
, and lands in flatter minima~\citep{izmailov2018averaging}. Across runs, we train a second
run at learning rate $5\times10^{-6}$ and average the pair, 
following the model-soup observation that same-initialization fine-tunes occupy one basin and
their average improves OOD accuracy at unchanged inference cost~\citep{wortsman2022modelsoups}.
On Whisper specifically, within-run averaging (MAST) improves 15.0 to 13.9 and across-run 
averaging (MAcT) to 13.2, a 12\% relative gain~\citep{ducorroy2025merging}, again 
dysarthric-speech, dev-subset numbers, which we treat only as an existence proof for acoustic-OOD Turkish. 

We did not evaluate trim-and-sign merging (TIES~\citep{yadav2023ties}): with a single task
vector its sign election is vacuous and it reduces to magnitude pruning, which keeps the
specialization we want to shrink. 

Model selection closes the loop. Candidates are scored on four axes (\sysname{tsc\_heldout},
FLEURS-TR test, \sysname{medv3\_val}, \sysname{tvoice\_heldout}), accepted only if they beat
raw \sysname{large-v3} on \sysname{tsc\_heldout} while conceding at most 0.3 points on
\sysname{tvoice\_heldout}, and ranked among the accepted by the mean of \sysname{tsc\_heldout}
and \sysname{medv3\_val}. FLEURS-TR test is scored and reported but does not enter the ranking,
which is what lets it serve as the out-of-distribution probe. 
inverting the in-domain selection that produced the narrowing. The executed champion
satisfies the pre-specified rule with room to spare: 17.01\% vs.\ 20.28\% on
\sysname{tsc\_heldout}, and it wins \sysname{tvoice\_heldout} outright (4.66\% vs.\ 6.32\%)
with nothing conceded (Table~\ref{tab:recovery}). 
Both of those axes are held out at the utterance level from sources that also feed training, so
they measure the rule being satisfied rather than generalization to an unseen distribution;
Section~\ref{sec:results} reports the independent axis separately and
Section~\ref{sec:limitations} states what the distinction costs. 


\section{TurkSpeak Bench}\label{sec:bench}

The failure mode documented here is not something the community's current instruments are
built to catch. Existing Turkish ASR leaderboards do not score controlled degradation: their
public views are clean read and broadcast speech, and the one noisy split is private and cannot
be re-run, whereas deployments encounter telephone lines, VoIP codecs and noisy rooms. 
The gap is not specific to Turkish. The Open ASR Leaderboard evaluates 86 systems across twelve
datasets and standardizes both text normalization and efficiency reporting, but its tracks
consist of clean short-form, multilingual and long-form speech. Its multilingual track covers
five European languages and no Turkish. Its authors also concede that test-set contamination
cannot be ruled out, and they list private evaluation sets as future
work \citep{srivastav2025openasr}. 
We therefore built TurkSpeak Bench, a robustness-first Turkish ASR leaderboard. Rather than
report a single clean-set score, it ranks models on how gracefully they degrade across
conditions \citep{turkspeakbench2026}.

Version~0 adopts the protocol of the cross-team validation reported in
Table~\ref{tab:fleurs}. It takes 100 utterances from the CC-BY-4.0 FLEURS-TR test
split \citep{conneau2023fleurs} and expands them into twenty conditions: clean audio,
MUSAN~\citep{snyder2015musan} noise at 20, 10 and 0~dB SNR, Opus re-encoding at 32, 16, 12 and
8~kbps, and the twelve noise$\times$codec combinations. 
Each model receives four aggregate rows, covering clean, $+$noise, $+$codec and
$+$noise$+$codec conditions. A narrowed model cannot hide behind a single clean
number. 
The harness deliberately matches the public board's backend and normalizer, so that the two
leaderboards remain directly comparable \citep{ysdede2026leaderboard}. Inference runs through
faster-whisper/CTranslate2 with beam~5 in fp16 on an RTX~5090, both reference and hypothesis
pass through \code{trnorm} normalization, and corpus-level WER and CER are computed
with \code{jiwer}. 
The interface is a fork of the ysdede leaderboard's MIT-licensed codebase, and every row lives
in a public, versioned data store (\repo{RsGoksel/turkspeak-bench-store}, MIT). 
The leaderboard is public; it was released on 2026-07-21 and its address is given under Data
and code availability. 

\section{Experimental Setup}\label{sec:setup}

\phead{Models and lineage}
Table~\ref{tab:lineage} maps every system in this paper onto its initialization, adapter
configuration, training data and model-selection signal. We use internal run names throughout.
\sysname{vh5\_\allowbreak 0d} is the public board entry \repo{RsGoksel\_\allowbreak ITU\_Mainframe},
and \sysname{vh5\_\allowbreak gen} is the in-domain push that followed
it. 
\sysname{robust\_\allowbreak v1} is a degradation-robust fine-tune of the same lineage, trained
with on-the-fly noise and reverberation augmentation (footnote~$b$ of Table~\ref{tab:lineage}
gives the exact recipe), \sysname{md\_\allowbreak v1} the multi-domain anti-forgetting run,
and \sysname{wiseft\_\allowbreak a0.5} an interpolation of two fine-tunes in weight space.

All fine-tunes are LoRA runs \citep{hu2022lora} whose adapters are merged into the base weights
after training. Each descendant then initializes from its parent's merged checkpoint. This gives
the lineage \sysname{vh5\_\allowbreak 0d}\,$\rightarrow$\,\sysname{robust\_\allowbreak v1}\,$\rightarrow$\,\sysname{md\_\allowbreak v1},
with \sysname{vh5\_\allowbreak gen} branching directly from \sysname{vh5\_\allowbreak 0d}. Every fine-tuned system evaluated
here descends from the board entry.
\sysname{wiseft\_\allowbreak a0.5} is the training-free midpoint of two fine-tunes in weight
space,
\[
  \theta \;=\; 0.5\,\theta_{\sysname{robust\_v1}} \;+\; 0.5\,\theta_{\sysname{vh5\_0d}},
\]
described further in \S\ref{sec:recovery-weights}. 
Unmodified \sysname{whisper-large-v3} and \sysname{whisper-large-v3-turbo} serve as zero-shot
baselines.

\phead{Training data and early stopping}
Full mixture compositions and provenance appear in Table~\ref{tab:data}, while
Table~\ref{tab:lineage} lists the headline volumes. \sysname{vh5\_\allowbreak 0d} trains on 997.2~h,
53.7\% of which comes from a single community podcast aggregate. 
\sysname{vh5\_\allowbreak gen} trains on 478.2~h, 53\% of it synthetic TurkishVoice
TTS. 
\sysname{robust\_\allowbreak v1} uses 662~h of clean audio, 
and \sysname{md\_\allowbreak v1} uses 864.8~h. 

The dev-set column encodes the design axis this paper turns on. \sysname{vh5\_\allowbreak 0d} stopped
on FLEURS-dev. \sysname{vh5\_\allowbreak gen} stopped instead on 2{,}000 held-out clips of its own majority source, the narrowest selection signal available, and trained for 12{,}000
steps at learning rate $10^{-5}$ with batch size 16. 
\sysname{md\_\allowbreak v1} stopped on a 938-clip mix of 600 TurkishVoice and 338 FLEURS-dev
utterances. 

One legacy defect deserves flagging. The \sysname{vh5\_\allowbreak 0d} corpus itself contained 8.3~h of
FLEURS, so its early-stop signal was not strictly held out. 
The recovery pipeline removes FLEURS from training entirely, reserving the dev split for early
stopping and the test split as an uncontaminated out-of-distribution probe. 

\phead{Chronology and advance specification}
Results are reported as of the 2026-07-21 cut-off. Four controls run later, each after the
reported protocol had been frozen: the seed replication and the full-epoch control of
Sections~\ref{sec:crosslingual} and~\ref{sec:results} on 28--29~July~2026, the per-utterance
re-decode behind the Table~\ref{tab:fleurs} intervals on 2026-08-14, and the decoder-fixed
comparison of Section~\ref{sec:results} on 2026-08-16.
The recovery methodology of Sections~\ref{sec:recovery} and~\ref{sec:results} was fixed before
any large-v3 checkpoint existed. The mixture build and the first training run took place on
2026-07-07, the weight-space stage on 2026-07-18, and all evaluations reported as of the cut-off were re-scored
under one faster-whisper protocol on 2026-07-21; the later controls state their own decoder
where it differs. The board maintainer's private-split re-score of the
recovery models was not obtainable, and no such number is claimed for them.

\phead{Recovery run}
The large-v3 recipe of Section~\ref{sec:recovery} was fixed before launch. It uses LoRA
$r{=}64$, $\alpha{=}128$ on $\{q,v\}$ at learning rate $10^{-5}$, with per-GPU batch 8 and
gradient accumulation 2 under DDP across two RTX~5090s, giving an effective batch of 32. Early stopping runs on FLEURS-dev, now genuinely out of mixture. 
As executed on 2026-07-07, an 8\,h wall-clock cap cut the planned 10{,}000-step budget to
6{,}000 steps. The best checkpoint fell at step 3{,}000, at a FLEURS-dev WER of 10.60\%, and
that checkpoint was the one merged and evaluated. 
Launch was gated on the final mixture build, 875.2~h as built, 
and on a hard contamination gate (\code{assert\_\allowbreak no\_overlap}) over dual
text-normalized exclusion sets. 
Post-training we apply three weight-space operations. The first averages the run's saved
checkpoints, three as executed. This is MAST-style trajectory averaging
\citep{ducorroy2025merging}, long standard in end-to-end ASR \citep{karita2019comparative} and
known to land in flatter minima \citep{izmailov2018averaging}. 
The second is an optional further run at learning rate $5\times10^{-6}$ whose endpoint is
averaged with the first, in the manner of MAcT-style across-run
averaging \citep{ducorroy2025merging} and of model soups \citep{wortsman2022modelsoups}. 
The third is a WiSE-FT sweep toward the raw base \citep{wortsman2022wiseft}
(Eq.~\eqref{eq:wiseft}) with $\alpha \in \{0.3, 0.4, 0.5, 0.6, 0.7\}$. Interpolating toward
$\theta_0$ itself is the methodological fix over \sysname{wiseft\_\allowbreak a0.5}. 
All three were executed and appear in Table~\ref{tab:recovery}: the WiSE sweep on 2026-07-07,
trajectory averaging and across-run souping on 2026-07-18. Trajectory averaging is the recipe's
final model. 

\phead{Model selection}
Candidates are scored on four axes and ranked on two of them, \sysname{tsc\_heldout} and
\sysname{medv3\_val}; FLEURS-TR test and \sysname{tvoice\_heldout} are reported but do not
enter the ranking. The primary proxy
is \sysname{tsc\_\allowbreak heldout}, comprising 5.0~h and 4{,}181 clips split off the
human-transcribed TSC corpus. Audio-level deduplication excludes it from the recovery mixture,
and it stands in for the board's private held-out distribution. 
Three further proxies complement it: FLEURS, \sysname{medv3\_\allowbreak val} (synthetic medical
TTS) and \sysname{tvoice\_\allowbreak heldout} (held-out TurkishVoice, the board's synthetic
in-domain set). 
The acceptance rule is fixed in advance (\S\ref{sec:recovery-weights}). 

\phead{Evaluation protocol}
All WER is computed after \code{trnorm} normalization of both reference and hypothesis. Three
evaluation surfaces are involved, and we state the aggregation of each explicitly. The cross-team
FLEURS-TR benchmark (Table~\ref{tab:fleurs}) and TurkSpeak~Bench both report corpus-level WER
computed with \code{jiwer}. 
The public board rows are produced end to end by the maintainer's WhisperX harness, in float16
with \code{condition\_\allowbreak on\_previous\_text} disabled on an NVIDIA L4, and are not
self-reported \citep{ysdede2026leaderboard}. 
The internal selection proxies of Table~\ref{tab:proxy} instead use mean-per-utterance
WER. 
Our own inference runs on faster-whisper/CTranslate2 with beam~5 in fp16, the same backend
family as the public board, which keeps backend and normalizer aligned across all three
surfaces. 
The FLEURS-TR dev split of 338 utterances serves exclusively for early stopping. The test split
of 715 utterances never enters any training mixture; it is scored but does not enter the selection ranking, which
is what lets it function as the
out-of-distribution probe. 
Internal evaluations and the TurkSpeak Bench harness run on RTX~5090 hardware; 
the recovery run trains on the two-GPU DDP configuration above. 


\section{Results}\label{sec:results}

\begin{figure}[t]
\sidecapoff
\centering
\includegraphics[width=0.86\linewidth]{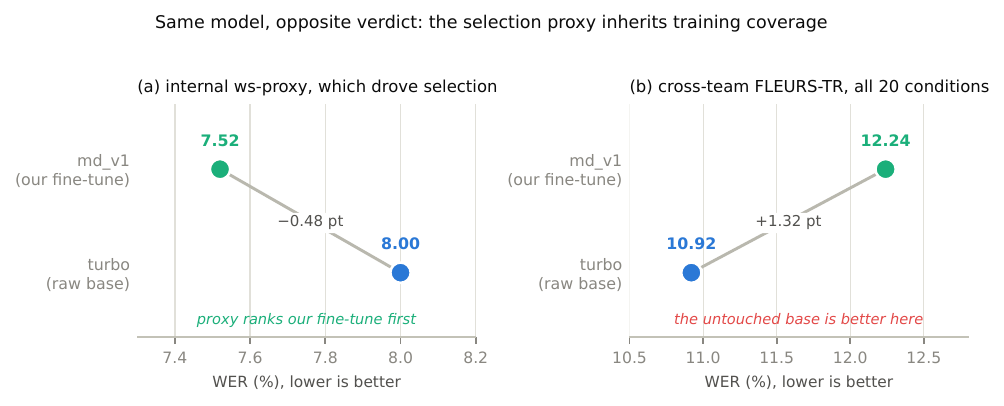}
\caption{The selection proxy conceals the narrowing it was meant to catch.
\emph{(a)}~On the internal three-axis suite the fine-tune ranks ahead of the raw base.
\emph{(b)}~On the cross-team FLEURS-TR grid the order reverses. Same models, same week,
opposite verdicts; the difference is where the evaluation audio came
from. 
}
\label{fig:proxy}
\end{figure}

\phead{Proxy blindness}
In-house proxies underestimate narrowing (Table~\ref{tab:proxy},
Figure~\ref{fig:proxy}). The 120-clip three-axis suite that drove turbo-generation model
selection ranked \sysname{md\_v1} above raw \sysname{large-v3} on the worldspeech-derived YS-0
proxy (7.52 vs.\ 7.79) and at near-parity on the medical axis (3.00 vs.\ 2.96), while external
evidence disagreed: on FLEURS-TR it trails even raw turbo by 1.32 points
ALL-20. 
(\sysname{md\_v1} was never scored on the public board. The submission repository moved to its
weights on 2026-06-29, but the board row was scored on 2026-06-27 and still reflects
\sysname{vh5\_0d}.) 
A suite that small, drawn from the training pools, is exactly the instrument a practitioner would
trust. It is also exactly the one that inverted the external verdict.

Scale alone does not fix it. The same inversion appears at $n{=}4{,}181$: on
\sysname{tsc\_heldout} the recovery fine-tune beats the raw base by 3.2 points while trailing it
on FLEURS-TR (Table~\ref{tab:recovery}): a same-provenance proxy saying ``better'' while the
independent axis does not confirm it, on a sample thirty-five times larger. 
This establishes something narrower than a causal claim about provenance: clip-level holdout
from a training-source distribution is \emph{insufficient} at either scale. The two proxies also
differ in speakers, acoustics and content, so provenance is not isolated here. Our recommendation is practical, and we claim no causal account: a selection suite should include at least one
distribution produced independently of the training pools.

We must be candid that our own recovery does not fully honour that rule. \sysname{tsc\_heldout}
is carved from the TSC corpus that also feeds the mixture. Its audio is deduplicated out of
training, so no clip is seen twice, but the recording conditions, speakers and collection
pipeline are shared. It is a held-out \emph{sample}, not an independent \emph{distribution}. We
used it because no Turkish corpus of independent provenance and comparable size was available to
us, and we report the consequence: the recovery's primary selection signal
satisfies the letter of held-out evaluation and not the principle this paper argues for. The
FLEURS-TR test set is the closest we come to it: excluded from training entirely and of
independent provenance from the recovery mixture, which makes it the strongest Turkish probe
available to us and the axis on which we report parity. It is not independent at the level of
the distribution, because its development split drove early stopping;
Section~\ref{sec:limitations} states what that costs the claim. 

\begin{figure}[t]
\sidecapoff
\centering
\includegraphics[width=0.82\linewidth]{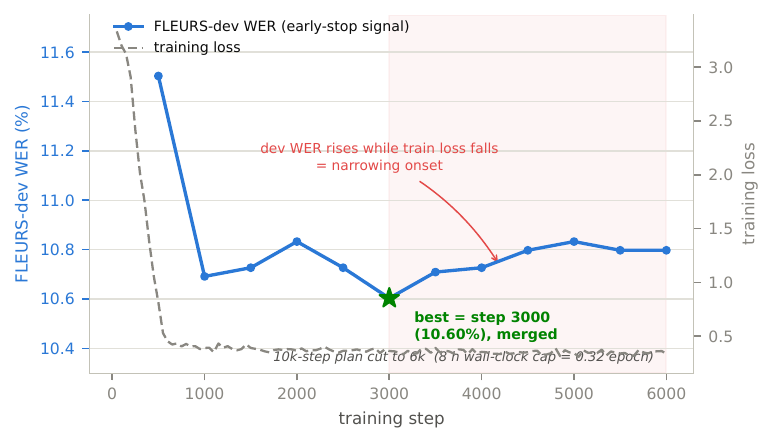}
\caption{Recovery run \sysname{lv3\_md} (large-v3, 875.2\,h mixture): FLEURS-dev WER
(early-stop signal, left axis) and training loss (right axis) over steps. The dev WER bottoms
at step 3{,}000 (10.60\%, merged checkpoint, green star) and creeps up while the training loss keeps falling: the overfitting signature within one run. The 10k-step plan was cut to 6{,}000 by
an 8\,h wall-clock cap (0.32 epoch). Data from the run's \code{trainer\_state}.} 
\label{fig:trainingcurve}
\end{figure}

The same pattern is already sitting, unnamed, in the published low-resource ASR literature.
A concurrent Kazakh study fine-tunes \sysname{whisper-large-v3-turbo} on mixtures of Common
Voice (CVC), FLEURS and a song corpus, and states its protocol plainly: ``Validation was
performed on the CVC and FLEURS validation sets,'' with early stopping at a patience of two
epochs~\citep{yeshpanov2026songs}. 
The third evaluation corpus, KSC2, exists only as a test split, so it never participates in
selection. 
Under that protocol the CVC+FLEURS configuration looks like the winner: it delivers the study's best FLEURS score (7.1 normalised WER), while on the corpus outside the selection
pools it moves from 9.3 to 13.8, a 48\% relative degradation against the same paper's
KSC2-trained reference point. 
The author reports the loss and names it ``forgetting''. However, the paper is likewise explicit that
one source ``only helps in-domain \dots\ but remains weak out-of-domain.'' 
Absent from that work is the step our results make: attributing the invisibility of that loss to the
\emph{selection instrument} itself. 
Read through the lens of this paper, its numbers corroborate the selection failure rather than replicate the inversion: the same-pool proxy favours a configuration that is weak on KSC2 in a second language, a different corpus ecosystem and by a different team, obtained before we looked, and free of our own analysis choices.\phead{The recovery measurement}
The acceptance rule of
Section~\ref{sec:recovery-weights} was fixed before any large-v3 checkpoint existed, so the
numbers below could not be curated after the fact. 
The 875.2\,h mixture was built on 2026-07-07 and \sysname{lv3\_md} was trained on it the same
day: 6{,}000 steps under an 8\,h cap, best checkpoint at step 3{,}000 with FLEURS-dev 10.60\%.
The training curve already previews the thesis. Dev WER bottoms at step 3{,}000 and rises from
there, while the training loss keeps falling (Figure~\ref{fig:trainingcurve}). 

That cap left the run at 0.32 epoch, which invites the objection that the result is
provisional. We tested it. Repeating the recipe for a full 18{,}526 steps and applying the same
trajectory averaging to both runs deepens the same-provenance gain: \sysname{tsc\_heldout} moves
from $-6.18$ to $-8.08$ against the base. Those two figures are measured on the greedy
transformers stack this control uses, where the base scores 22.40 and the 0.32-epoch model
16.22; Table~\ref{tab:recovery} reports the same pair on faster-whisper at beam~5, where the
base scores 20.28 and the gain is 3.27. The stack, not the model, accounts for the difference:
it costs the base 2.12 points and gains the fine-tune 0.79. Longer exposure deepens the narrowing on the axis that
shares provenance with the training mixture, so the cap did not weaken the
diagnosis. 

\phead{Separating the decoder from the budget}
That control decodes greedily, whereas Table~\ref{tab:recovery} uses beam~5, so budget and
decoder moved together in it. Reading a budget effect out of two model-versus-base intervals
would also be the wrong test: a difference of significances is not the significance of a
difference. We therefore re-ran base, the 0.32-epoch average and the full-epoch average on the
same 715 utterances with the decoder held fixed. 
Three things follow, all under paired bootstrap ($B{=}10{,}000$, per-utterance).
First, the parity of Table~\ref{tab:recovery} replicates in a second inference stack. Both
measurements use beam~5 and the same 715 utterances; they differ in the serving stack, Hugging
Face \code{generate} here against faster-whisper/CTranslate2 there. The 0.32-epoch model sits
$+0.21$ WER points from the base with $[-0.05,+0.47]$, where Table~\ref{tab:recovery} measured
$-0.10$ with $[-0.20,+0.40]$. Both intervals span zero, so we report parity and not either point
estimate. 
Second, the budget contrast itself is not significant: full epoch against 0.32 epoch is $+0.09$
with $[-0.13,+0.32]$. Longer training does not measurably widen the FLEURS-TR gap once the
decoder is fixed, and we withdraw that reading of the greedy control. 
Third, the decoder is a real effect for both models. Moving from beam~5 to greedy costs the base
$+0.22$ $[+0.10,+0.35]$ and the full-epoch model $+0.37$ $[+0.15,+0.59]$. The point estimate is
larger for the fine-tune, but the paired difference between the two penalties is $+0.15$ with
$[-0.08,+0.38]$, so we do not claim that the decoder penalises one model more than the other.
What the control does establish is that decoder choice moved the FLEURS-TR numbers by more than
the training budget did, which is what inflated the apparent budget effect. 
The narrowing diagnosis is unaffected by any of this, because it rests on gaps of whole points
rather than tenths. What the experiment does establish is narrower and worth stating plainly:
FLEURS-TR parity survives a change of inference stack, and the deepening we can demonstrate with
longer training is on the same-provenance axis, not on the independent one. 

\begin{figure}[t]
\sidecapoff
\centering
\includegraphics[width=0.86\linewidth]{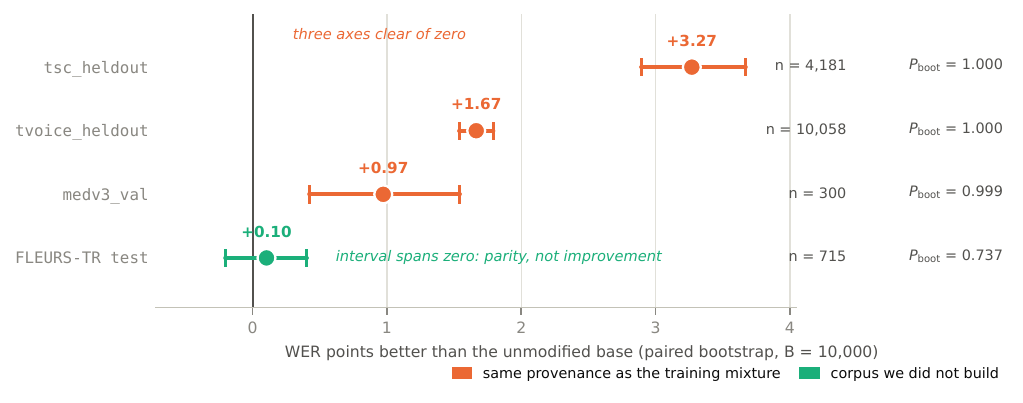}
\caption{Effect size and uncertainty for the recovery model on all four selection axes, against
unmodified \sysname{whisper-large-v3}. Positive means fewer errors. Intervals are paired
bootstrap ($B{=}10{,}000$, per-utterance); $P_{\mathrm{boot}}$ is the proportion of resamples the
recovery model wins, not a null-hypothesis $p$-value. Colour encodes provenance: the three axes
clearing zero share it with the training mixture, the axis spanning zero does
not. 
}
\label{fig:recoveryci}
\end{figure}

The recipe's final model comes from the weight-space stage executed
on 2026-07-18: uniform trajectory averaging (MAST~\citep{ducorroy2025merging};
SWA~\citep{izmailov2018averaging}) over the run's three saved checkpoints. Against the raw base
it wins the three same-provenance axes decisively: \sysname{tsc\_heldout} $+3.27$
($\mathrm{CI}_{95}=[+2.89,+3.67]$), \sysname{tvoice} $+1.67$ ($[+1.54,+1.79]$), \sysname{medv3} $+0.97$ ($[+0.42,+1.54]$), and reaches parity on FLEURS-TR, $+0.10$ with
$[-0.20,+0.40]$ spanning zero (Table~\ref{tab:recovery},
Figure~\ref{fig:recoveryci}). Signs here follow the figure convention, where positive means
fewer errors than the base; Table~\ref{tab:recovery} reports raw WER, so the same cell appears
there as 8.16 against 8.26. No axis shows a statistically established regression. Every WiSE
interpolation degrades monotonically toward the base on \sysname{tsc\_heldout}, so the champion
under the pre-specified proxy is the averaged fine-tune. The sweep's own numbers sit outside
Table~\ref{tab:recovery} because they were scored on the first 800 \sysname{tsc}/\sysname{tvoice}
clips (best point $\alpha^{*}{=}0.7$: 19.10 and 5.48 there, 8.32 FLEURS-TR, 3.94
\sysname{medv3}). 

An internal comparison keeps us from over-reading this. On our own bench, whose FLEURS-TR view
scores the 100-utterance degradation subset, the champion reads 5.91\% clean against the base's
6.37\%; on the full 715-utterance test the same pair reverses. 
The verdict flips with the evaluation sample, so we treat the full test as the honest probe,
claim no clean-FLEURS win, and report the bench subset only for continuity with the cross-team
validation. 

\phead{The synthetic-share ablation}
We put the mixture's most questionable ingredient to the test. Synthetic
TTS is 29.0\% of the recovery mixture, which invites the objection that a robustness recipe
is partly built on machine-generated prosody. 
We therefore re-ran the identical recipe (same base, adapter, learning rate, step budget) on a
mixture with every synthetic source removed (464{,}580 clips, 21.6\% smaller) and evaluated it under the same protocol. 
Removing synthetic speech leaves clean read speech and spontaneous YouTube untouched
(FLEURS-TR 8.33 vs.\ 8.34; \sysname{tsc\_heldout} 16.98 vs.\ 17.06) while costing 0.38 points
on medical speech (4.26 vs.\ 3.88) and 1.26 points on the synthetic-domain axis (6.07 vs.\
4.81). 
The ablation speaks to two questions, within what single-run point estimates can show. The synthetic share is far from inert ballast. It buys measurable accuracy on two axes, and the feared prosodic over-fit leaves no trace in the point estimates: if synthetic data were
narrowing the model toward machine prosody, the synthetic-free model would read clean speech
better, and it does not. 
The averaged model also improves on the best single checkpoint everywhere, so the averaging
stage costs nothing at inference and dominates checkpoint selection. On FLEURS it reaches 8.16
against 8.34, and the one near-tie is \sysname{tsc\_heldout} at 17.01 against
17.06. 
The ablation rows sharpen the mechanism. A second run at learning rate $5{\times}10^{-6}$
under the same budget stays under-trained at 0.32 epoch, reaching FLEURS 8.25 but
\sysname{tsc} 19.56. Souping it with run-1, in the manner of
MAcT~\citep{ducorroy2025merging}, \emph{dilutes} rather than helps: FLEURS reaches 8.29 and
every cell falls behind MAST. Under a fixed wall-clock budget, averaging \emph{within} the
converged run therefore beats averaging \emph{across} runs whenever the second run cannot be
trained to comparable quality. This is a budget-regime complement to the dysarthric-speech
setting in which across-run averaging won~\citep{ducorroy2025merging}. 
The public-board re-score that alone can settle YS-0 remains with the maintainer. 

\begin{table}[t]
\centering
\caption{Recovery ablation (corpus WER, \%; fine-tune and WiSE sweep measured 2026-07-07, the
weight-space stage 2026-07-18). Rows are the recovery candidates of
Section~\ref{sec:recovery}; columns are the composite selection-proxy suite, all scored under
one harness. 
Harness: faster-whisper/CTranslate2 float16, beam 5, no VAD. Every cell was re-scored on
2026-07-18 over the full sets, so the columns are composition-matched: $N$ = 4{,}181
\sysname{tsc\_heldout}, 715 FLEURS-TR test, 300 \sysname{medv3\_val}, 10{,}058
\sysname{tvoice\_heldout}. The WiSE-$\alpha$ sweep is reported separately in the text because it
was measured on 2026-07-07 over the first 800 \sysname{tsc}/\sysname{tvoice} clips and would not
be comparable in these columns. 
The WiSE-$\alpha$ sweep is reported in the text (best point $\alpha^{*}{=}0.7$); it degrades
monotonically toward the base on \sysname{tsc\_heldout}. 
Acceptance rule, fixed in advance: beat raw \sysname{large-v3} on \sysname{tsc\_heldout} while conceding $\leq$0.3 points on \sysname{tvoice\_heldout}, satisfied by the champion (bold). 
YS-0 re-scoring is an external dependency on the leaderboard maintainer. 
}
\label{tab:recovery}
\footnotesize
\setlength{\tabcolsep}{4pt}%
\begin{tabular}{@{}L{0.33\linewidth}cccc@{}}
\toprule
Candidate & \sysname{tsc\_heldout} & FLEURS-TR test & \sysname{medv3\_val} & \sysname{tvoice\_heldout} \\
\midrule
\sysname{whisper-large-v3} (raw)          & 20.28 & 8.26 & 4.68 & 6.32 \\ 
\sysname{lv3\_md} (large-v3 FT, best ckpt) & 17.06 & 8.34 & 3.88 & 4.81 \\ 
\sysname{lv3\_md2} (2nd run, lr $5{\times}10^{-6}$) & 19.56 & 8.25 & 4.40 & 6.23 \\ 
\quad + across-run soup (MAcT)            & 19.47 & 8.29 & 4.06 & 5.89 \\ 
\midrule
\textit{ablation:} synthetic-free mixture & 16.98 & 8.33 & 4.26 & 6.07 \\ 
\textbf{\sysname{lv3\_md} + trajectory averaging (MAST), ours} & \textbf{17.01} & \textbf{8.16} & \textbf{3.71} & \textbf{4.66} \\ 
\bottomrule
\end{tabular}
\end{table}

\begin{table}[t]
\centering
\caption{Internal three-axis selection proxies (mean-per-utterance WER, \%; $N{=}120$ clips per axis; 
2026-06-29, turbo generation). 
Axes: medical validation, MediaSpeech-clean, and a worldspeech-derived YS-0 proxy
(ws-proxy); \sysname{mdwise-0.7/-0.5} and \sysname{wiseft-champion} are internal
weight-interpolation selection candidates. Bold = best per column. These are small-$N$ selection metrics; they do not serve as benchmarks: the ws-proxy ranked \sysname{md\_v1}
above raw \sysname{large-v3}, 
a signal contradicted by the independent FLEURS-TR evidence (Table~\ref{tab:fleurs}) 
and never tested on the board. 
}
\label{tab:proxy}
\small
\begin{tabular}{@{}lccc@{}}
\toprule
Model & Medical & Clean & ws-proxy \\
\midrule
\sysname{md\_v1}                        & 3.00          & 10.98          & \textbf{7.52} \\ 
\sysname{whisper-large-v3} (raw)        & \textbf{2.96} & 12.22          & 7.79          \\ 
\sysname{mdwise-0.7}                    & 3.07          & 10.00          & 7.98          \\ 
\sysname{whisper-large-v3-turbo} (raw)  & 3.41          & 12.68          & 8.00          \\ 
\sysname{wiseft-champion}               & 3.40          & 9.82           & 8.29          \\ 
\sysname{mdwise-0.5}                    & 3.43          & \textbf{9.80}  & 8.36          \\ 
\bottomrule
\end{tabular}
\end{table}

We summarise the state of the work as of the 2026-07-21 cut-off. The final mixture is built,
comprising 592{,}836 clips and 875.2\,h assembled on 2026-07-07. 
The \sysname{lv3\_md} large-v3 fine-tune is trained, merged and selected as champion under the
pre-specified rule. 
Table~\ref{tab:recovery} reports the full weight-space stage, covering the WiSE-$\alpha$ sweep,
MAST-style trajectory averaging, which is the recipe's final and best model, and MAcT-style
across-run averaging \citep{ducorroy2025merging}. 
Two matters lie outside our reach. The recovery candidates were not re-scored on the board's
private split, which only its maintainer can run, so no such number is claimed for
them. 
TurkSpeak Bench, with its eighteen scored models, is public, and its address appears under
Data and code availability. 

\section{How far the effect travels: language and architecture}\label{sec:crosslingual}

Everything above rests on a single language, corpus ecosystem and team. To test whether proxy
blindness is a Turkish accident or a property of the selection protocol itself, we
replicated the design in Kazakh, choosing it because it satisfies the one condition the
argument needs: three test corpora of \emph{independent provenance}, none of which we built.

\phead{Design} The setup mirrors the Turkish case exactly. We LoRA fine-tune
\sysname{whisper-large-v3} (rank 64, $\alpha=128$, \code{q/v\_proj}) on a deliberately narrow, clean pool (Common Voice~kk train plus FLEURS-kk train, 3{,}794 utterances /
12.51\,h) for 1{,}200 steps at an effective batch of 32. Checkpoint selection uses a
\emph{same-pool} proxy: the validation splits of those same two corpora (876 utterances).
The out-of-pool probe is KSC2-test (9{,}350 utterances / 15.9\,h), which contributes nothing
to training or selection and carries six labelled origin domains. A contamination gate over
text-normalized transcripts found one overlapping utterance in KSC2, which we removed. WER
uses a language-neutral normalizer. Our Turkish normalizer would have mangled Kazakh
numerals. Scores are corpus-level WER (total errors over total reference words) under greedy
decoding with the HuggingFace generate path, identical for base and fine-tune. This differs
from the beam-5 CTranslate2 backend used for the Turkish tables, so only
within-language contrasts are read here. 

\phead{Pipeline validation} Before comparing anything we checked the ruler: our
zero-shot \sysname{large-v3} scores 33.19 WER on FLEURS-kk, against 33.1 reported
independently for the same model and split~\citep{yeshpanov2026songs}. The measurement is on
the same scale as the published literature. 

\phead{Result} Table~\ref{tab:kk} shows the inversion, with paired bootstrap intervals
($B=10{,}000$, utterance-level resampling). On the pools the proxy is drawn from, the fine-tune looks like a win: FLEURS-kk improves by 1.29 points with an interval excluding
zero, Common Voice improves directionally. On the probe the proxy never sees, it
\emph{loses} by 1.25 points, and that interval excludes zero as well. A practitioner
following the standard protocol would ship this model and record it as an improvement. 

\begin{table}[t]\centering\small
\caption{Kazakh replication. The same-pool proxy ranks the narrow fine-tune above the base;
the out-of-pool probe reverses the verdict. Paired bootstrap, $B=10{,}000$.}
\label{tab:kk}
\begin{tabular}{llrrrc}
\toprule
Evaluation set & Pool & base & narrow FT & $\Delta$ & 95\% CI \\
\midrule
Common Voice~kk test & in  & 48.89 & 48.02 & $-0.86$ & $[-2.92,+1.21]$ \\
FLEURS-kk test       & in  & 33.19 & \textbf{31.90} & $\mathbf{-1.29}$ & $[-2.20,-0.43]$ \\
\midrule
KSC2 test            & \textbf{out} & \textbf{57.39} & 58.64 & $\mathbf{+1.25}$ & $[+0.62,+1.87]$ \\
\midrule
\multicolumn{6}{l}{\emph{KSC2 by origin domain (all out-of-pool)}}\\
\quad parliament   & out & 85.36 & 79.70 & $-5.66$ & $[-10.35,-1.19]$ \\
\quad crowdsourced & out & 46.08 & 47.16 & $+1.08$ & $[+0.43,+1.71]$ \\
\quad tv\_news     & out & 52.42 & 53.72 & $+1.30$ & $[+0.52,+2.04]$ \\
\quad radio        & out & 66.44 & 68.07 & $+1.63$ & $[-0.76,+3.66]$ \\
\quad talkshow     & out & 61.99 & 64.93 & $+2.94$ & $[+0.55,+5.00]$ \\
\quad podcasts     & out & 64.14 & 68.18 & $+4.04$ & $[+2.78,+5.24]$ \\
\bottomrule
\end{tabular}
\end{table}

\phead{How the damage is distributed} The domain breakdown is the part we
did not have in Turkish, and it points at the mechanism. The fine-tuning pool is read speech:
FLEURS is read Wikipedia, Common Voice is volunteers reading prompts. Among the six out-of-pool
domains (Figure~\ref{fig:kazakh}), the ordering follows acoustic distance from that pool:
conversational podcasts degrade most ($+4.04$), and the only domain that moves the other way is
parliament ($-5.66$), the most read-like of the six: formal, prepared, single-speaker. 

Because this breakdown involves six simultaneous comparisons, we report it under
Bonferroni correction as well, and we let the correction discipline the claim. Three
degradations survive it individually (99.17\% intervals: podcasts $[+2.30,+5.61]$, tv\_news
$[+0.30,+2.29]$, crowdsourced $[+0.19,+1.95]$); talkshow and radio do not, and
\emph{neither does parliament's improvement} $[-11.93,+0.81]$, which rests on 773
utterances. We therefore claim the ordering as directional evidence for the mechanism and
only the three degradations as individually established. The paper's central contrast is
unaffected: correcting the three axis-level tests, FLEURS-kk remains
$[-2.39,-0.22]$ and KSC2 remains $[+0.48,+2.02]$, so the proxy-versus-probe inversion
survives correction on both sides. 

The term itself needs a clarification. The five held-out KSC2 domains differ from the pool
acoustically, but also in register, spontaneity, topic and speaker population. Our own proposed
mechanism is a decoder specializing to the pool's linguistic surface. That is not a purely
acoustic story. We therefore use \emph{acoustic narrowing} to denote contraction over
acoustic-domain \emph{bundles} as they occur in practice, not over an isolated acoustic
variable. Separating the two would require adaptation and testing on matched text under
different recording conditions. We did not run that, and where the distinction matters we say
domain rather than acoustic. 

With that reading, the model did not get worse at Kazakh. It got better at the conditions it was shown and worse at the ones it was not. That is what we mean by acoustic narrowing, now observed
in a second language, on corpora we did not build, with the direction of every effect fixed in
advance by the Turkish case. 

The same conclusion is now supported from two independent directions: the controlled replication
here, and the re-reading of a concurrent Kazakh study in Section~\ref{sec:results} whose
published numbers show the same inversion under the same same-pool selection
protocol~\citep{yeshpanov2026songs}. 

\phead{Architecture: which system carries the narrowing} Both
replications so far adapt an attention encoder--decoder, so we designed a controlled test of
whether proxy blindness follows the \emph{protocol} or the \emph{architecture}. A naive comparison would leave two things free that must be held fixed. Adaptation capacity: our initial attempt tuned only MMS's language adapter, 0.24\% of parameters against Whisper's 2.00\% under
LoRA, and the model barely moved. We discarded it and applied LoRA to MMS at the same rank and
target modules, giving 1.60\%. The second, less obvious, is base coverage: \sysname{MMS-1B-all} was pretrained on FLEURS, so a design that fine-tunes on FLEURS cannot separate architecture from
the base already knowing the pool. 

We removed that confound by construction. Pool \emph{and} probe now
come from a single corpus, KSC2, which appears in neither model's pretraining. The pool is its
crowdsourced training domain, 5{,}840 utterances over 12.50\,h in one recording condition, and so
genuinely narrow. The same-pool proxy is a held-out slice of that domain, 979 utterances over
2.00\,h. The out-of-pool probe is the corpus's \emph{other five} origin domains, 6{,}022
utterances over 8.92\,h. Corpus, transcription standard and collection
pipeline are therefore held constant, while the held-out origin domains differ as real-world acoustic-domain bundles, the object the phenomenon is named after. A contamination gate
found no overlap, and pool, proxy and probe are pairwise disjoint by audio path. 

\begin{table}[t]\centering\small
\caption{Controlled architecture test. Same pool, same probe, same corpus, matched adaptation
capacity, no pretraining-coverage confound. Out-of-pool WER change after fine-tuning; positive
is worse. Paired bootstrap $B=10{,}000$; domain rows use 99\% intervals (Bonferroni, five
tests). Every cell excludes zero. The bracketed values are paired-bootstrap intervals over
evaluation samples; the seed row instead reports training-run variability across seeds 42/43/44
under identical settings. Domain rows come from the seed-42 run.} 
\label{tab:arch}
\begin{tabular}{lrcrc}
\toprule
& \multicolumn{2}{c}{Whisper (attn.\ enc--dec, 2.00\%)} & \multicolumn{2}{c}{MMS (CTC enc-only, 1.60\%)}\\
\cmidrule(lr){2-3}\cmidrule(lr){4-5}
Probe & $\Delta$ & 95/99\% CI & $\Delta$ & 95/99\% CI \\
\midrule
\textbf{All five held-out domains} & $\mathbf{+2.23}$ & $[+1.19,+3.26]$ & $\mathbf{-3.46}$ & $[-3.81,-3.12]$\\
\quad \emph{mean $\pm$ SD over 3 seeds} & $+2.28$ & $\pm 0.08$ & $-3.60$ & $\pm 0.19$\\
\midrule
\quad parliament (prepared, formal) & $-10.02$ & $[-16.60,-3.66]$ & $-3.52$ & $[-5.01,-2.13]$\\
\quad tv\_news                      & $+1.76$  & $[+0.38,+3.19]$  & $-4.44$ & $[-5.23,-3.66]$\\
\quad radio                         & $+3.56$  & $[+0.27,+6.51]$  & $-3.43$ & $[-4.62,-2.28]$\\
\quad talkshow                      & $+6.58$  & $[+2.70,+10.41]$ & $-1.49$ & $[-3.02,-0.10]$\\
\quad podcasts (spontaneous)        & $+6.72$  & $[+4.61,+8.82]$  & $-2.93$ & $[-3.64,-2.20]$\\
\bottomrule
\end{tabular}
\end{table}

The two architectures move in opposite directions, and both intervals exclude zero
(Table~\ref{tab:arch}). Fine-tuning the attention encoder--decoder on one acoustic domain costs
2.23 points across the five it did not see. Fine-tuning the CTC model on the identical data
\emph{gains} 3.46. The domain rows sharpen this. Whisper reproduces the acoustic-distance
ordering of the Kazakh grid above, and this time it survives Bonferroni correction:
against a read-style crowdsourced pool, the most read-like held-out domain (prepared, formal, single-speaker parliament speech) improves by 10.02 points, while spontaneous
conversational podcasts lose 6.72. MMS shows no such ordering at all. All five domains improve,
including the two furthest from the pool. 

\phead{Not an artifact of our metric} A reader may reasonably ask whether a result computed
with our own harness survives a different measuring stick. We recomputed the same comparison
from the same stored per-utterance outputs under three independent metrics: corpus WER (what we
report), mean-per-utterance WER (what the leaderboards use, weighting every utterance equally),
and corpus CER, which is indifferent to word boundaries and to Kazakh's agglutinative
morphology. All six intervals exclude zero and every one agrees in sign: Whisper degrades by
$+2.23$, $+2.20$ and $+7.63$ points respectively, MMS improves by $-3.46$, $-3.79$ and $-1.98$.
The character-level effect for Whisper is the largest of the three. The damage therefore lies in the model emitting different characters, consistent with a decoder that has specialized to the pool's
linguistic surface. 

We state the outcome at the level the design supports: under this protocol the effect appeared
in one architecture and reversed in the other, which makes narrowing a property of the system under adaptation. It does not identify \emph{which} architectural property
carries it. The two systems differ in pretraining data and objective, tokenization, output
representation, parameter count and headroom, and matching trainable-parameter ratio (2.00\%
vs.\ 1.60\%) equalizes budget but not adaptation dynamics. A decoder-frozen versus
encoder-frozen ablation within one model family, and an RNN-T arm (autoregressive but not attention-based) are the interventions that would isolate the variable. We did not run
them. Read as a two-system comparison, the result still carries a mechanism. Whisper's decoder is an autoregressive language model, free to
specialize to the text, register and delivery of whatever pool it is shown; CTC is
frame-synchronous and has no such component, so it exposes no comparable part to the same
pressure; whether its encoder contracts is something this design does not isolate. On this
reading acoustic narrowing would be a property of fine-tuning a model whose decoder can overfit the pool's linguistic surface, which is the family that currently dominates low-resource ASR leaderboards, and precisely where the
selection warning of this paper applies. 

\section{Discussion and Limitations}\label{sec:limitations}

\begin{figure}[t]
\sidecapoff
\centering
\includegraphics[width=0.86\linewidth]{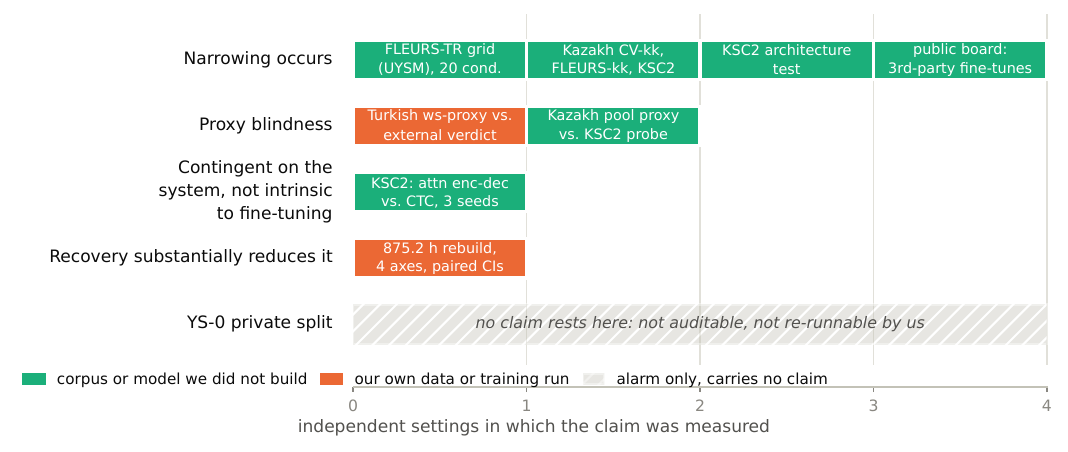}
\caption{What each claim rests on. A block is one setting in which the claim was \emph{measured}
independently: a distinct corpus, team or model family. YS-0 is drawn at full width but carries
no weight; the deletion test below shows every claim survives without it. 
}
\label{fig:claims}
\end{figure}

Figure~\ref{fig:claims} states the shape of the evidence before the caveats: which claim was
measured in how many independent settings, who built the corpora and models behind each, and
where nothing is claimed at all. The limitations below are the reasons those bars are the
length they are.

\phead{Held out at the utterance level, and not at the distribution level}
This is the most consequential limitation for the recovery claim. FLEURS-TR test is held out at the
utterance level but not at the distribution level. The recovery run early-stops on FLEURS-dev,
and dev and test share corpus construction, speaking style, recording protocol and
transcription convention. Calling that split ``fully excluded'' is accurate about
training data and misleading about model selection, and we retire the phrase. The direction of
that leak matters: a development split drawn from the same corpus biases the test split
\emph{towards} the model it early-stopped, so it works in favour of a win. We claim parity, and
the conservative reading survives the confound. The 8.16 versus 8.26 comparison supports parity on a held-out split of a distribution that \emph{did} steer early stopping. It says nothing about recovery on an unseen distribution. The genuinely
independent Turkish evidence in this paper is the cross-team FLEURS-TR degradation grid, which
was built by another group under a cross-team protocol and which motivated rather than selected
the recipe. The
Kazakh and architecture experiments of Section~\ref{sec:crosslingual} carry the
distribution-independent load. Closing this properly needs a Turkish corpus untouched by
training, early stopping, proxy construction and recipe design, ideally scored externally. We
did not have one. 

\phead{The scope of ``unseen''}
We cannot audit YS-0's composition, and its owner's description (roughly 18 hours of noisy real-world recordings) plus the public tooling that
appears to have produced it are the only signals available; any profile we give is
inference. 
This is precisely why the paper's evidential weight sits elsewhere. A reviewer is entitled to
ask what would remain if YS-0 were removed from the paper entirely, and the answer is: the
FLEURS-TR inversion under twenty controlled conditions, on a grid another team built from public
data. The Kazakh replication on corpora built by neither of us. The controlled architecture
comparison; and the recovery measurement on four axes whose per-utterance outputs we release.
The diagnosis, the proxy-blindness result and the recovery all survive that deletion. Losing it would cost the anecdote that prompted the work. That is the correct role for a split
we can neither inspect nor re-score.

\phead{Training variance}
Bootstrap intervals resample test utterances, not runs, so we
repeated both architecture arms at three seeds (Table~\ref{tab:arch}). The sign is identical in
all six runs and the across-seed SD is an order of magnitude below the effect. The recovery run
itself remains a single trajectory and we claim no seed-level stability for it. 

\phead{External validity, and what remains of it}
The diagnosis now rests on two languages,
four test sets and three teams: the Turkish case, the independent FLEURS-TR benchmark, and the
Kazakh replication of Section~\ref{sec:crosslingual} on corpora we did not build, plus a concurrent Kazakh study whose own published numbers show the same
inversion~\citep{yeshpanov2026songs}. 
Architecture dependence is measured, and it is now a result. The design gives pool and probe one
corpus that neither model saw in pretraining, and matches adaptation capacity. Under it the
attention encoder--decoder loses 2.23 points out-of-pool while the CTC encoder-only model gains
3.46, every domain-level interval excluding zero
(Table~\ref{tab:arch}). Breadth is what remains open. This is one architecture pair in one language, and the mechanism we infer, that an autoregressive decoder is the component that specializes, is a reading of the pattern, not something we
manipulated directly. An RNN-T model, which is autoregressive but not attention-based, is the
experiment that would separate \emph{autoregressive} from \emph{attention}. We did not run it. A second caution is metric-level: agglutinative morphology makes word-level
error harsh relative to character-level error~\citep{mussakhojayeva2023multilingual}, so absolute
WER values should be compared within a language, not across the Turkish and Kazakh tables; only the \emph{signs} and intervals are meant to be read together.
Scope needs two narrower qualifications. The first is deployment form. Our LoRA adapters are
merged into the served weights, so the findings characterize merged-LoRA deployment rather than
adapter-switched serving \citep{qian2024learn}.
The second is provenance, uneven across the mixture. YODAS ``manual'' captions are
user-uploaded rather than certified human transcripts \citep{li2023yodas}, and the
diagnosis-phase podcast aggregate carried no verifiable aggregate licence. That is precisely why
the recovery mixture rebalances toward CC0 Common Voice~22, CC-BY-3.0 YODAS and MIT-carded TSC
(Table~\ref{tab:data}). 

\section{Conclusion}\label{sec:conclusion}

A Turkish ASR fine-tune won a public leaderboard under third-party scoring and lost, on the
board's private held-out split, to the unmodified base it was built from. The inversion
reproduced on another team's twenty-condition benchmark. 
We named the failure mode acoustic narrowing and isolated its portable corollary: the in-house
proxy meant to catch it concealed it instead, so reference-distribution accuracy cannot police
this regime and a selection proxy must be held out from an independently produced
distribution. 

We then fixed the claim's boundaries by measurement. Replicating the design in Kazakh reproduced
the inversion on corpora we did not build, and the damage fell in order of acoustic distance from the training pool. That is the mechanism the name asserts. 
A controlled architecture test, holding corpus, pool and adaptation capacity fixed, found the
narrowing in the attention encoder--decoder while the CTC encoder-only model improved. Whether narrowing occurs therefore depends on the system being adapted. Which architectural property carries the effect remains a hypothesis, since the two models
differ in more than one respect. 

The recovery direction was fixed in advance and executed in full: an 875.2-hour mixture whose
provenance is audited and disclosed per source, two components of it unverified, a higher-capacity base, and base-referenced selection with trajectory
averaging. At the reported operating point it substantially reduces the narrowing, with decisive
margins on three same-provenance held-out axes and parity on FLEURS-TR. Within that run,
trajectory averaging gave the strongest FLEURS-TR result and the best joint standing across the
evaluated axes. The full-epoch control places the boundary: trained longer, the same recipe deepens the
same-provenance gain without measurably widening the FLEURS-TR gap, once the decoder is held
fixed. We do not attribute the improvement over the earlier
turbo fine-tunes to averaging: capacity, mixture, adapter configuration and selection signal
all changed together. 

\begin{figure}[tb]
\sidecapoff
\centering
\includegraphics[width=0.86\linewidth]{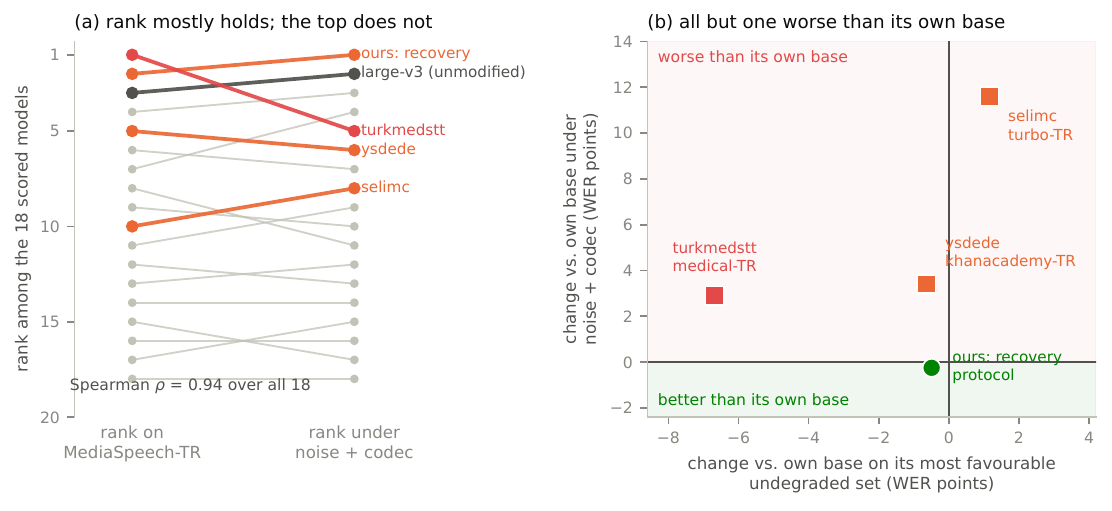}
\caption{The released benchmark, read through this paper's question. \emph{(a)}~The eighteen
scored systems ranked by MediaSpeech-TR and again under FLEURS-TR noise$+$codec.
\emph{(b)}~The four entries whose base model is documented, by the publisher's model card
(squares) or by our fact ledger (circles), plotted as change from that base; the horizontal axis
is each model's most favourable undegraded set, excluding any the board flags as
in-domain-contaminated for it. Excluded from the pairing are a distillation, a model declaring
no base, \sysname{vh5\_gen}, and systems trained from scratch. Board WER uses the board's own
normalizer and decoding, so only within-board differences are read here. 
}
\label{fig:bench_close}
\end{figure}

TurkSpeak Bench turns the lesson into infrastructure, and its public rows let the question be
put to models we did not build (Figure~\ref{fig:bench_close}). Ranking the eighteen scored
systems by a clean in-domain set and again under noise-plus-codec leaves the global order largely
intact, at $\rho = 0.94$. What moves is the top, the only part of a table anyone acts on. The in-domain leader falls to fifth and the unmodified \sysname{large-v3} rises to second. The best-of-five axis in panel~(b) is chosen to give every fine-tune its strongest
case, and the contamination exclusion behind it runs against us: the flag can only be set where
training data is known, so it fires on our rows and not on third-party ones. Dropping it moves
no point by more than 0.2\,WER. Among the four entries whose base model is documented, three are
worse than that base under degradation, two of them while beating it on at least one undegraded set. The one
exception is the model built under the recovery protocol of Section~\ref{sec:recovery}, and its margins there are small, 0.49 and 0.25 WER points, which is the parity of Table~\ref{tab:recovery} restated on public infrastructure. We did
not control what those other teams trained on and the board flags several of these rows for
in-domain contamination. They extend the pattern without establishing it, and they do show that it is not ours alone. For low-resource languages, where a single community board
can steer an entire modeling ecosystem, we call for leaderboards that score robustness (degraded, codec-distorted, and genuinely held-out conditions) so that the next win does not
narrow the next model.

\appendix
\setlength{\textfloatsep}{10pt}
\setlength{\intextsep}{10pt}
\section{Supporting tables and figures}
\label{sec:appendix}

Every claim in this paper is stated and evidenced in the main text. The material below repeats
those numbers in a second view, or lists configuration detail a reader reproducing the work will
want, and is placed here so the argument reads without interruption.

\begin{table}[H]
\centering
\caption{The seven boards of the ysdede Turkish ASR
leaderboard~\citep{ysdede2026leaderboard}: WER of our entry \sysname{vh5\_0d}
against the two strongest unmodified Whisper checkpoints, as of the 2026-06-27 store update, re-verified 2026-07-04. 
$\Delta=$ ours $-$ large-v3 in WER points (negative favors the fine-tune);
rank is our position among all models listed on that board. Rows sorted by
$\Delta$.}
\label{tab:board}
\small
\begin{tabular}{@{}llrrrrc@{}}
\toprule
 & & \multicolumn{3}{c}{WER (\%)} & & \\
\cmidrule(lr){3-5}
Board & Composition & ours & large-v3 & turbo & $\Delta$ & rank \\
\midrule
MediaSpeech            & public; media speech                  &  9.02 & 11.93 & 12.17            & $-2.91$ & \#1/16 \\ 
cv17-fixed (test)      & public; read speech                   &  8.73 &  9.53 & n/a$^{\dagger}$  & $-0.80$ & \#1/11 \\ 
Average                & aggregate; a model's own rows$^{\ddagger}$ & 9.71 &  9.90 & 10.62$^{\dagger}$ & $-0.19$ & \#1/17 \\ 
tr-med-audio           & private; medical                      &  6.05 &  6.03 &  6.18            & $+0.02$ & \#3/13 \\ 
Unseen                 & aggregate; the two private sets       &  9.44 &  8.69 &  9.02            & $+0.75$ & \#3/13 \\ 
TurkishVoice           & public; synthetic TTS                 & 11.92 & 10.67 & 11.54            & $+1.25$ & \#4/16 \\ 
Yeni Split 0 (YS-0)    & private; noisy real-world             & 12.82 & 11.35 & 11.87            & $+1.47$ & \#4/16 \\ 
\bottomrule
\end{tabular}

\smallskip
\begin{minipage}{0.93\textwidth}
\footnotesize
$^{\dagger}$\,turbo has no cv17-fixed \emph{test} row in the benchmark store, so its five-set Average is not composition-matched to ours and large-v3's and is shown for context only. 
$^{\ddagger}$\,the Average board averages whatever rows each model has, so the set varies; ours and large-v3 cover the same five sets. 
\end{minipage}
\end{table}

\begin{table}[H]
\centering
\caption{Model lineage and configurations. All fine-tunes are LoRA runs merged into the base
weights after training; descendants initialize from their parent's merged checkpoint. The final
row is the executed large-v3 recovery run of Section~\ref{sec:recovery}. \code{n/a} marks
fields that do not apply (training-free interpolation) or are not in our verified run registry.}
\label{tab:lineage}
\footnotesize
\setlength{\tabcolsep}{3.5pt}
\begin{tabular}{@{}L{0.191\linewidth}L{0.169\linewidth}L{0.129\linewidth}r L{0.198\linewidth} r@{}}
\toprule
Model (role) & Base / init & LoRA $r/\alpha$, targets & Data (h) & Dev set / early stop & Steps \\
\midrule
\sysname{vh5\_\allowbreak 0d}\newline (board entry) & \sysname{whisper-\allowbreak large-v3-\allowbreak turbo} & $32/64$, $\{q,v\}$ & 997.2 & FLEURS-dev & n/a$^{a}$ \\[2pt] 
\sysname{vh5\_\allowbreak gen}\newline (in-domain push) & \sysname{vh5\_\allowbreak 0d} (merged) & $64/128$, $\{q,k,v,o\}$ & 478.2 & 2{,}000 TurkishVoice clips (in-domain) & 12k \\[2pt] 
\sysname{robust\_\allowbreak v1}\newline (degradation-robust)$^{b}$ & \sysname{vh5\_\allowbreak 0d} (merged) & $64/128$, $\{q,k,v,o\}$ & 662 & n/a & 15k \\[2pt] 
\sysname{md\_\allowbreak v1}\newline (multi-domain)$^{c}$ & \sysname{robust\_\allowbreak v1} (merged) & $32/64$, $\{q,k,v,o\}$ & 864.8 & 600 TurkishVoice $+$ 338 FLEURS-dev (mixed) & 8k \\[2pt] 
\sysname{wiseft\_\allowbreak a0.5}\newline (FT$\leftrightarrow$FT interp.) & $0.5\,\theta_{\sysname{robust\_v1}}$ $+\;0.5\,\theta_{\sysname{vh5\_0d}}$ & n/a & n/a & n/a & n/a \\[2pt] 
\sysname{lv3\_\allowbreak md}\newline (recovery, 2026-07-07)$^{d}$ & \sysname{whisper-\allowbreak large-v3} & $64/128$, $\{q,v\}$ & 875.2 & FLEURS-dev (OOD) & 6k \\ 
\bottomrule
\end{tabular}

\vspace{3pt}
\begin{minipage}{\linewidth}\footnotesize
$^{a}$Stopped by FLEURS-dev early stopping under the training script's default schedule; step
budget not registered. 
\\
$^{b}$On-the-fly degradation augmentation applied with probability $p{=}0.45$: one of additive
MUSAN noise \citep{snyder2015musan} at 0--20\,dB SNR, additive DEMAND noise
\citep{thiemann2013demand} at the same range, or convolution with a simulated room impulse
response, sampled uniformly; clean audio otherwise. This is the complete recipe for the row. 
\\
$^{c}$On-the-fly degradation augmentation, $p{=}0.40$. 
\\
$^{d}$Planned at 10k steps; executed at 6{,}000 under an 8\,h wall-clock cap ($=0.32$ epoch);
best checkpoint at step 3{,}000 by FLEURS-dev WER (10.60\%). 
\end{minipage}
\end{table}

\begin{figure}[H]
\sidecapoff
  \centering
  \includegraphics[width=0.72\linewidth]{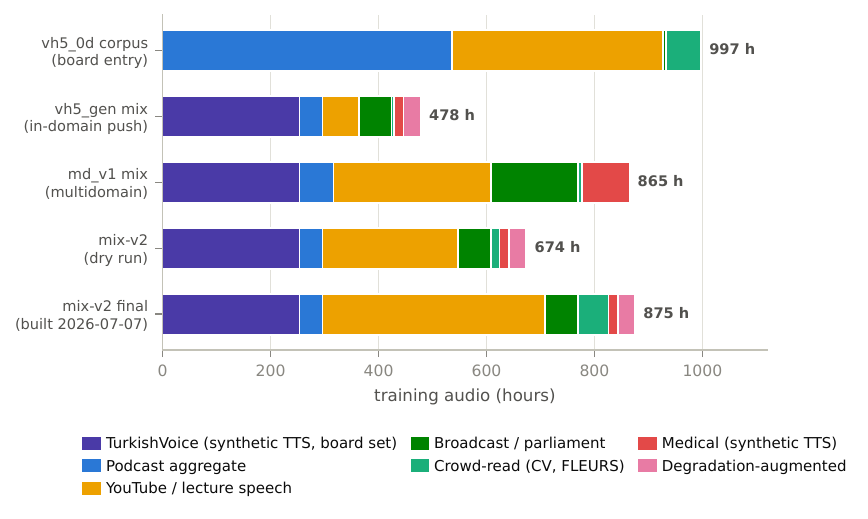}
  \caption{Evolution of the training mixture across model generations. The board entry
  \sysname{vh5\_0d} trained on 997.2\,h dominated by a 53.7\% podcast aggregate; 
  the in-domain push \sysname{vh5\_gen} shrank the mix to 478.2\,h with a 53\% synthetic-TTS 
  share; 
  \sysname{md\_v1} re-diversified to 864.8\,h; 
  \sysname{mix-v2} (dry run, 673.6\,h) re-centres on human- and user-transcribed spontaneous 
  YouTube speech (TSC, YODAS \sysname{tr000}) and CC0 read speech (CV22). 
  The final build, executed 2026-07-07, measures 875.2\,h. 
  Synthetic TTS components are labelled as such throughout.}
  \label{fig:mix}
\end{figure}

\begin{figure}[H]
\sidecapoff
  \centering
  \includegraphics[width=0.72\linewidth]{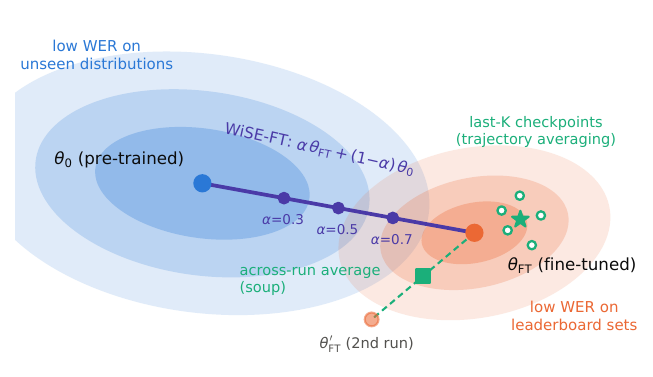}
  \caption{Weight-space schematic of the recovery toolkit. Fine-tuning carries the
  pre-trained weights $\theta_0$, whose basin covers unseen acoustic distributions, to
  $\theta_\mathrm{FT}$, whose basin covers the leaderboard sets. WiSE-FT
  (Eq.~\eqref{eq:wiseft}) re-approaches the base along the linearly connected path; trajectory
  averaging merges the last $K$ checkpoints of one run; across-run averaging merges
  independently trained runs. Only the interpolation path returns toward $\theta_0$, the anchor our earlier fine-tune-to-fine-tune interpolation \sysname{wiseft\_a0.5} lacked. 
  }
  \label{fig:weightspace}
\end{figure}

\begin{figure}[H]
\sidecapoff
\centering
\includegraphics[width=0.72\linewidth]{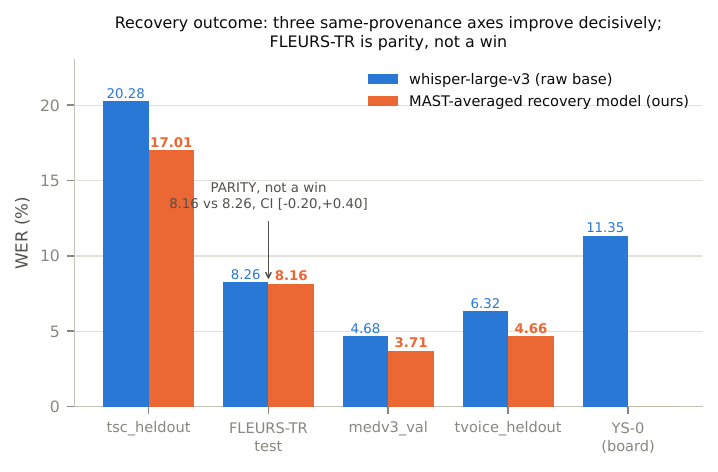}
\caption{Recovery outcome (Table~\ref{tab:recovery}): the recipe's final model, the trajectory-averaged fine-tune, vs.\ raw \sysname{whisper-large-v3} across the selection-proxy
suite. The averaged model wins the three same-provenance axes by clear margins and reaches \emph{parity} on FLEURS-TR (8.16 vs.\ 8.26; the paired interval
$[-0.20,+0.40]$ includes zero). We label that cell parity throughout; a 0.10-point difference
with an interval spanning zero is not a win.} 
\label{fig:recovery}
\end{figure}

\begin{figure}[H]
\sidecapoff
\centering
\includegraphics[width=0.72\linewidth]{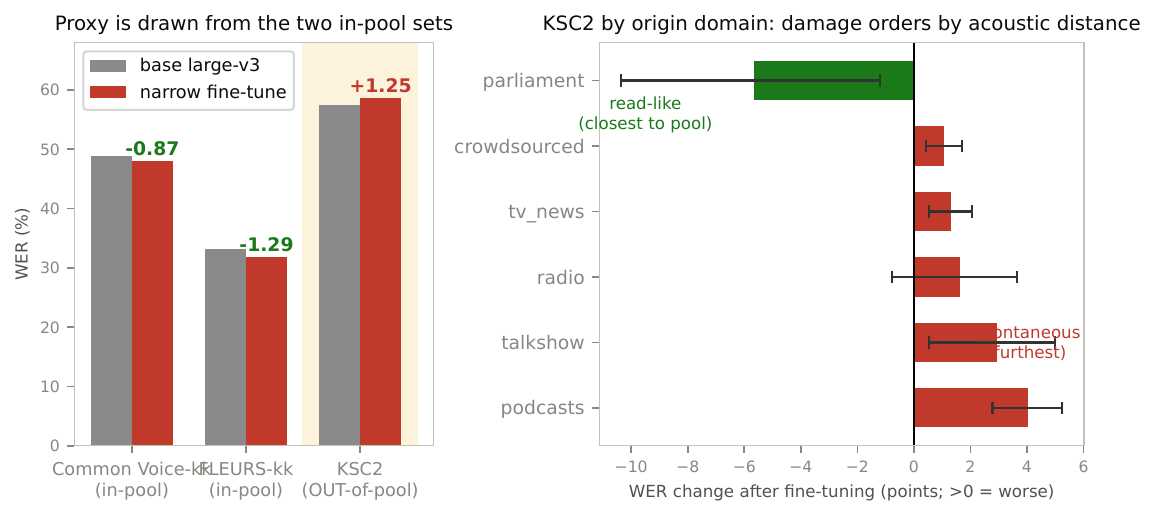}
\caption{Kazakh replication. \emph{Left:} the fine-tune improves on both pools the selection
proxy is drawn from and loses on the probe of independent provenance (shaded): the same inversion as the Turkish case, in a corpus ecosystem we did not build. \emph{Right:} within
that probe the damage is not diffuse. Ordering the six origin domains by outcome recovers an
acoustic-distance ordering: read-like parliament speech, closest to the read-speech training
pool, is the only domain that moves the other way, while spontaneous conversational podcasts
fall furthest. Bars are WER change in points (positive = worse); whiskers are 95\%
paired-bootstrap intervals, $B=10{,}000$. Under Bonferroni correction for the six domains,
podcasts, tv\_news and crowdsourced remain individually significant; talkshow, radio and
parliament do not (Section text).} 
\label{fig:kazakh}
\end{figure}

\section*{Data and code availability}
The models the paper's claims rest on are released under a single Hugging Face namespace,
\url{https://huggingface.co/RsGoksel}. Table~\ref{tab:artifacts} maps the identifier used in the
text to the corresponding repository. Four intermediate checkpoints of the lineage in
Table~\ref{tab:fleurs} (\sysname{md\_v1}, \sysname{robust\_v1}, \sysname{wiseft\_a0.5},
\sysname{vh5\_gen}) and the two controls of Section~\ref{sec:results} (the synthetic-free
mixture and the full-epoch run) are not released as weights. Their per-utterance outputs,
manifests and decoding configurations are in the code repository, so every interval reported for
them can be recomputed without the weights.

\begin{table}[H]
\centering\small
\begin{tabular}{@{}L{3.7cm}L{5.6cm}L{4.9cm}@{}}
\toprule
\textbf{Identifier in text} & \textbf{Repository} & \textbf{Description} \\
\midrule
\sysname{vh5\_0d}          & \repo{RsGoksel\_ITU\_\allowbreak Mainframe}
                           & the narrowed fine-tune of Section~\ref{sec:diagnosis} \\
\sysname{vh5\_0d}, for inference & \repo{\dots Mainframe-ct2}
                           & the same weights, converted \\
\sysname{lv3\_md}, averaged & \repo{turkspeak-lv3-\allowbreak ITU\_Mainframe}
                           & the recovery model of Section~\ref{sec:recovery} \\
TurkSpeak Bench            & \repo{spaces/\dots\allowbreak turkish-\allowbreak asr-\allowbreak benchmark-stt}
                           & the leaderboard, live \\
\bottomrule
\end{tabular}
\caption{Model identifiers used in the text and their corresponding repositories.}
\label{tab:artifacts}
\end{table}

\noindent
Three further resources accompany this work. Code for training, evaluation and statistical
analysis is released at \url{https://github.com/RsGoksel/acoustic-narrowing}, along with the per-utterance
outputs underlying every reported confidence interval and a ledger linking each number in this
paper to the artifact that produced it. The data store supporting the leaderboard is available
at \url{https://huggingface.co/datasets/RsGoksel/turkspeak-bench-store}. The Kazakh replication
draws exclusively on public corpora (Common Voice~kk, FLEURS-kk and KSC2); its manifests,
contamination gate and per-utterance results are included in the same repository. One resource
cannot be released: the private split YS-0 remains under the control of the leaderboard
maintainer.

\noindent
One licensing consequence follows from the mixture audit of Section~\ref{sec:recovery}. The
recovery mixture's largest single component, 29.0\% of the final build, is synthetic speech
carried under CC-BY-NC-4.0. The released recovery weights therefore inherit a non-commercial
restriction, and their model card states it. A further 42.2 hours, 4.8\% of the build, is a
community podcast aggregate whose licence we could not verify; Table~\ref{tab:data} marks it as
such. The diagnosis-phase checkpoints are released for reproduction of the reported evaluations
rather than for deployment.

\section*{Acknowledgments}
We thank the maintainer of the ysdede Turkish ASR leaderboard for independently scoring our
submissions on the private held-out sets, and the UYSM team for constructing the FLEURS-TR
degradation benchmark, running its own reference models, cross-validating our evaluation
pipeline, and providing the computing hardware on which this work was carried out. UYSM is a
research unit at the author's institution; Section~\ref{sec:validation} states which parts of
the evaluation each group performed.

\section*{Declaration of AI use in the writing process}
In preparing this work the author used Claude (Anthropic) as a writing and engineering
assistant: for language editing, for restructuring and rewriting prose, for adversarial critique
of the argument across several rounds, and for producing the plotting and typesetting code behind
the figures and tables. The research itself was not delegated. The experimental design, the
training runs, the evaluation protocol and every reported number are the author's own, and each
number is traceable to the artifact that produced it through the fact ledger released with the
code. The author reviewed and edited all AI-assisted text, checked every claim against that
ledger, and takes full responsibility for the content of this publication.

\bibliographystyle{IEEEtranN}
\bibliography{refs}

\begin{thebibliography}{33}
\providecommand{\natexlab}[1]{#1}
\providecommand{\url}[1]{#1}
\csname url@samestyle\endcsname
\providecommand{\newblock}{\relax}
\providecommand{\bibinfo}[2]{#2}
\providecommand{\BIBentrySTDinterwordspacing}{\spaceskip=0pt\relax}
\providecommand{\BIBentryALTinterwordstretchfactor}{4}
\providecommand{\BIBentryALTinterwordspacing}{\spaceskip=\fontdimen2\font plus
\BIBentryALTinterwordstretchfactor\fontdimen3\font minus
  \fontdimen4\font\relax}
\providecommand{\BIBforeignlanguage}[2]{{%
\expandafter\ifx\csname l@#1\endcsname\relax
\typeout{** WARNING: IEEEtranN.bst: No hyphenation pattern has been}%
\typeout{** loaded for the language `#1'. Using the pattern for}%
\typeout{** the default language instead.}%
\else
\language=\csname l@#1\endcsname
\fi
#2}}
\providecommand{\BIBdecl}{\relax}
\BIBdecl

\bibitem[Srivastav et~al.(2025)Srivastav, Zheng, Bezzam, Le~Bihan, Koluguri,
  {\.Z}elasko, Majumdar, Moumen, and Gandhi]{srivastav2025openasr}
V.~Srivastav, S.~Zheng, E.~Bezzam, E.~Le~Bihan, N.~R. Koluguri, P.~{\.Z}elasko,
  S.~Majumdar, A.~Moumen, and S.~Gandhi, ``Open {ASR} leaderboard: Towards
  reproducible and transparent multilingual and long-form speech recognition
  evaluation,'' \emph{arXiv preprint arXiv:2510.06961}, 2025.

\bibitem[Gandhi et~al.(2022)Gandhi, von Platen, and Rush]{gandhi2022esb}
S.~Gandhi, P.~von Platen, and A.~M. Rush, ``{ESB}: A benchmark for multi-domain
  end-to-end speech recognition,'' \emph{arXiv preprint arXiv:2210.13352},
  2022.

\bibitem[{ysdede}(2026)]{ysdede2026leaderboard}
{ysdede}, ``{T}urkish {ASR} leaderboard,'' Hugging Face Space
  \url{https://huggingface.co/spaces/ysdede/turkish_asr_leaderboard}; benchmark
  data store \url{https://huggingface.co/datasets/ysdede/asr_benchmark_store},
  2026, community-run public benchmark; accessed 2026-07-04.

\bibitem[Radford et~al.(2023)Radford, Kim, Xu, Brockman, McLeavey, and
  Sutskever]{radford2023whisper}
A.~Radford, J.~W. Kim, T.~Xu, G.~Brockman, C.~McLeavey, and I.~Sutskever,
  ``Robust speech recognition via large-scale weak supervision,'' in
  \emph{Proceedings of the 40th International Conference on Machine Learning
  (ICML)}, ser. Proceedings of Machine Learning Research, vol. 202.\hskip 1em
  plus 0.5em minus 0.4em\relax PMLR, 2023, pp. 28\,492--28\,518,
  arXiv:2212.04356.

\bibitem[{OpenAI}(2024)]{openai2024turbo}
{OpenAI}, ``{W}hisper large-v3-turbo,'' Hugging Face model card
  \url{https://huggingface.co/openai/whisper-large-v3-turbo}, 2024, 809M
  parameters; number of decoding layers reduced from 32 to 4 relative to
  {W}hisper large-v3 (per model card, which links
  \url{https://github.com/openai/whisper/discussions/2363}); accessed
  2026-07-04.

\bibitem[Hu et~al.(2022)Hu, Shen, Wallis, Allen-Zhu, Li, Wang, Wang, and
  Chen]{hu2022lora}
E.~J. Hu, Y.~Shen, P.~Wallis, Z.~Allen-Zhu, Y.~Li, S.~Wang, L.~Wang, and
  W.~Chen, ``{LoRA}: Low-rank adaptation of large language models,'' in
  \emph{International Conference on Learning Representations (ICLR)}, 2022,
  arXiv:2106.09685.

\bibitem[Kumar et~al.(2022)Kumar, Raghunathan, Jones, Ma, and
  Liang]{kumar2022finetuning}
A.~Kumar, A.~Raghunathan, R.~Jones, T.~Ma, and P.~Liang, ``Fine-tuning can
  distort pretrained features and underperform out-of-distribution,'' in
  \emph{International Conference on Learning Representations (ICLR)}, 2022,
  arXiv:2202.10054.

\bibitem[Qian et~al.(2024)Qian, Tang, Ma, Knill, and Gales]{qian2024learn}
M.~Qian, S.~Tang, R.~Ma, K.~M. Knill, and M.~J.~F. Gales, ``Learn and don't
  forget: Adding a new language to {ASR} foundation models,'' in \emph{Proc.
  Interspeech 2024}, 2024, pp. 2544--2548, arXiv:2407.06800.

\bibitem[Carvalho et~al.(2026)Carvalho, Teixeira, Rolland, and
  Abad]{carvalho2026merging}
C.~Carvalho, F.~Teixeira, T.~Rolland, and A.~Abad, ``Exploring the potential
  and limitations of model merging for multi-domain adaptation in {ASR},''
  \emph{arXiv preprint arXiv:2603.05354}, 2026, submitted to INTERSPEECH 2026.

\bibitem[Conneau et~al.(2023)Conneau, Ma, Khanuja, Zhang, Axelrod, Dalmia,
  Riesa, Rivera, and Bapna]{conneau2023fleurs}
A.~Conneau, M.~Ma, S.~Khanuja, Y.~Zhang, V.~Axelrod, S.~Dalmia, J.~Riesa,
  C.~Rivera, and A.~Bapna, ``{FLEURS}: Few-shot learning evaluation of
  universal representations of speech,'' in \emph{2022 IEEE Spoken Language
  Technology Workshop (SLT)}.\hskip 1em plus 0.5em minus 0.4em\relax Doha,
  Qatar: IEEE, 2023, pp. 798--805.

\bibitem[Wortsman et~al.(2022{\natexlab{a}})Wortsman, Ilharco, Kim, Li,
  Kornblith, Roelofs, Gontijo-Lopes, Hajishirzi, Farhadi, Namkoong, and
  Schmidt]{wortsman2022wiseft}
M.~Wortsman, G.~Ilharco, J.~W. Kim, M.~Li, S.~Kornblith, R.~Roelofs,
  R.~Gontijo-Lopes, H.~Hajishirzi, A.~Farhadi, H.~Namkoong, and L.~Schmidt,
  ``Robust fine-tuning of zero-shot models,'' in \emph{Proceedings of the
  IEEE/CVF Conference on Computer Vision and Pattern Recognition (CVPR)}, 2022,
  pp. 7959--7971, arXiv:2109.01903.

\bibitem[Wortsman et~al.(2022{\natexlab{b}})Wortsman, Ilharco, Gadre, Roelofs,
  Gontijo-Lopes, Morcos, Namkoong, Farhadi, Carmon, Kornblith, and
  Schmidt]{wortsman2022modelsoups}
M.~Wortsman, G.~Ilharco, S.~Y. Gadre, R.~Roelofs, R.~Gontijo-Lopes, A.~S.
  Morcos, H.~Namkoong, A.~Farhadi, Y.~Carmon, S.~Kornblith, and L.~Schmidt,
  ``Model soups: averaging weights of multiple fine-tuned models improves
  accuracy without increasing inference time,'' in \emph{Proceedings of the
  39th International Conference on Machine Learning (ICML)}, ser. Proceedings
  of Machine Learning Research, vol. 162.\hskip 1em plus 0.5em minus
  0.4em\relax PMLR, 2022, pp. 23\,965--23\,998.

\bibitem[Yeshpanov(2026)]{yeshpanov2026songs}
R.~Yeshpanov, ``Using songs to improve kazakh automatic speech recognition,''
  \emph{arXiv preprint arXiv:2603.00961}, 2026, v3, 7 March 2026.

\bibitem[Recht et~al.(2019)Recht, Roelofs, Schmidt, and
  Shankar]{recht2019imagenet}
B.~Recht, R.~Roelofs, L.~Schmidt, and V.~Shankar, ``Do {ImageNet} classifiers
  generalize to {ImageNet}?'' in \emph{Proceedings of the 36th International
  Conference on Machine Learning (ICML)}, ser. Proceedings of Machine Learning
  Research, vol.~97.\hskip 1em plus 0.5em minus 0.4em\relax PMLR, 2019, pp.
  5389--5400, arXiv:1902.10811.

\bibitem[Tseng et~al.(2025)Tseng, Parcollet, van Dalen, Zhang, and
  Bhattacharya]{tseng2025contamination}
Y.~Tseng, T.~Parcollet, R.~van Dalen, S.~Zhang, and S.~Bhattacharya,
  ``Evaluation of {LLM}s in speech is often flawed: Test set contamination in
  large language models for speech recognition,'' \emph{arXiv preprint
  arXiv:2505.22251}, 2025.

\bibitem[Akeret(2026)]{akeret2026swissgerman}
F.~Akeret, ``Subtitle-aligned fine-tuning of {W}hisper for {S}wiss {G}erman
  {ASR}: Benchmark contamination, convention mismatch, and an honest baseline
  at 25.6\% {WER} (13.8\% {cWER}),'' \emph{arXiv preprint arXiv:2606.07608},
  2026.

\bibitem[Izmailov et~al.(2018)Izmailov, Podoprikhin, Garipov, Vetrov, and
  Wilson]{izmailov2018averaging}
P.~Izmailov, D.~Podoprikhin, T.~Garipov, D.~Vetrov, and A.~G. Wilson,
  ``Averaging weights leads to wider optima and better generalization,'' in
  \emph{Proceedings of the 34th Conference on Uncertainty in Artificial
  Intelligence (UAI)}, 2018, pp. 876--885, arXiv:1803.05407.

\bibitem[Karita et~al.(2019)Karita, Chen, Hayashi, Hori, Inaguma, Jiang,
  Someki, Soplin, Yamamoto, Wang, Watanabe, Yoshimura, and
  Zhang]{karita2019comparative}
S.~Karita, N.~Chen, T.~Hayashi, T.~Hori, H.~Inaguma, Z.~Jiang, M.~Someki,
  N.~E.~Y. Soplin, R.~Yamamoto, X.~Wang, S.~Watanabe, T.~Yoshimura, and
  W.~Zhang, ``A comparative study on transformer vs {RNN} in speech
  applications,'' in \emph{IEEE Automatic Speech Recognition and Understanding
  Workshop (ASRU)}, 2019, pp. 449--456, arXiv:1909.06317.

\bibitem[Ilharco et~al.(2023)Ilharco, Ribeiro, Wortsman, Gururangan, Schmidt,
  Hajishirzi, and Farhadi]{ilharco2023editing}
G.~Ilharco, M.~T. Ribeiro, M.~Wortsman, S.~Gururangan, L.~Schmidt,
  H.~Hajishirzi, and A.~Farhadi, ``Editing models with task arithmetic,'' in
  \emph{The Eleventh International Conference on Learning Representations
  (ICLR)}, 2023, arXiv:2212.04089.

\bibitem[Ducorroy and Riad(2025)]{ducorroy2025merging}
A.~Ducorroy and R.~Riad, ``Robust fine-tuning of speech recognition models via
  model merging: application to disordered speech,'' in \emph{Proc. Interspeech
  2025}, 2025, pp. 3279--3283, arXiv:2505.20477.

\bibitem[Mussakhojayeva et~al.(2023)Mussakhojayeva, Dauletbek, Yeshpanov, and
  Varol]{mussakhojayeva2023multilingual}
S.~Mussakhojayeva, K.~Dauletbek, R.~Yeshpanov, and H.~A. Varol, ``Multilingual
  speech recognition for {T}urkic languages,'' \emph{Information}, vol.~14,
  no.~2, p.~74, 2023.

\bibitem[Mercan et~al.(2023)Mercan, {\c{C}}epni, Ta{\c{s}}ar, and
  Ozan]{mercan2023performance}
{\"O}.~B. Mercan, S.~{\c{C}}epni, D.~E. Ta{\c{s}}ar, and {\c{S}}.~Ozan,
  ``Performance comparison of pre-trained models for speech-to-text in
  {T}urkish: {W}hisper-{S}mall and {W}av2{V}ec2-{XLS-R}-300{M},'' \emph{arXiv
  preprint arXiv:2307.04765}, 2023, in Turkish with English abstract.

\bibitem[Polat et~al.(2024)Polat, Turan, Ko{\c{c}}ak, and
  Ula{\c{s}}]{polat2024whisper}
H.~Polat, A.~K. Turan, C.~Ko{\c{c}}ak, and H.~B. Ula{\c{s}}, ``Implementation
  of a {W}hisper architecture-based {T}urkish automatic speech recognition
  ({ASR}) system and evaluation of the effect of fine-tuning with a low-rank
  adaptation ({LoRA}) adapter on its performance,'' \emph{Electronics},
  vol.~13, no.~21, p. 4227, 2024.

\bibitem[Kolobov et~al.(2021)Kolobov, Okhapkina, Omelchishina, Platunov,
  Bedyakin, Moshkin, Menshikov, and Mikhaylovskiy]{kolobov2021mediaspeech}
R.~Kolobov, O.~Okhapkina, O.~Omelchishina, A.~Platunov, R.~Bedyakin,
  V.~Moshkin, D.~Menshikov, and N.~Mikhaylovskiy, ``{MediaSpeech}:
  Multilanguage {ASR} benchmark and dataset,'' \emph{arXiv preprint
  arXiv:2103.16193}, 2021.

\bibitem[Ardila et~al.(2020)Ardila, Branson, Davis, Henretty, Kohler, Meyer,
  Morais, Saunders, Tyers, and Weber]{ardila2020commonvoice}
\BIBentryALTinterwordspacing
R.~Ardila, M.~Branson, K.~Davis, M.~Henretty, M.~Kohler, J.~Meyer, R.~Morais,
  L.~Saunders, F.~M. Tyers, and G.~Weber, ``{Common Voice}: A
  massively-multilingual speech corpus,'' in \emph{Proceedings of the Twelfth
  Language Resources and Evaluation Conference (LREC 2020)}.\hskip 1em plus
  0.5em minus 0.4em\relax Marseille, France: European Language Resources
  Association, 2020, pp. 4218--4222. [Online]. Available:
  \url{https://aclanthology.org/2020.lrec-1.520}
\BIBentrySTDinterwordspacing

\bibitem[{ysdede}(2024)]{asrtk2024}
{ysdede}, ``{ASRTK},'' GitHub repository \url{https://github.com/ysdede/asrtk},
  2024, open-source {P}ython toolkit designed to streamline the development and
  enhancement of {ASR} systems; MIT license; accessed 2026-07-04.

\bibitem[{erenfazlioglu}(2024)]{turkishvoice2024}
{erenfazlioglu}, ``{T}urkish neural voice dataset,'' Hugging Face dataset
  \url{https://huggingface.co/datasets/erenfazlioglu/turkishvoicedataset},
  2024, synthetic speech generated with Microsoft Text-to-Speech services;
  license CC-BY-NC-4.0; accessed 2026-07-04.

\bibitem[Snyder et~al.(2015)Snyder, Chen, and Povey]{snyder2015musan}
D.~Snyder, G.~Chen, and D.~Povey, ``{MUSAN}: A music, speech, and noise
  corpus,'' \emph{arXiv preprint arXiv:1510.08484}, 2015.

\bibitem[Li et~al.(2023)Li, Takamichi, Saeki, Chen, Shiota, and
  Watanabe]{li2023yodas}
X.~Li, S.~Takamichi, T.~Saeki, W.~Chen, S.~Shiota, and S.~Watanabe, ``{YODAS}:
  {Y}outube-oriented dataset for audio and speech,'' in \emph{2023 IEEE
  Automatic Speech Recognition and Understanding Workshop (ASRU)}.\hskip 1em
  plus 0.5em minus 0.4em\relax IEEE, 2023, pp. 1--8, arXiv:2406.00899.

\bibitem[{Mozilla Foundation}(2025)]{commonvoice22}
{Mozilla Foundation}, ``{Common Voice} corpus 22.0,'' Release
  cv-corpus-22.0-2025-06-20, \url{https://commonvoice.mozilla.org/en/datasets};
  community mirror
  \url{https://huggingface.co/datasets/fsicoli/common_voice_22_0}, 2025,
  license CC0-1.0; accessed 2026-07-04.

\bibitem[Yadav et~al.(2023)Yadav, Tam, Choshen, Raffel, and
  Bansal]{yadav2023ties}
P.~Yadav, D.~Tam, L.~Choshen, C.~Raffel, and M.~Bansal, ``{TIES}-merging:
  Resolving interference when merging models,'' in \emph{Advances in Neural
  Information Processing Systems (NeurIPS)}, vol.~36, 2023, arXiv:2306.01708.

\bibitem[G{\"u}nd{\"u}z(2026)]{turkspeakbench2026}
K.~G. G{\"u}nd{\"u}z, ``{TurkSpeak} bench: A robustness-first {T}urkish {ASR}
  leaderboard,'' Hugging Face Space
  \url{https://huggingface.co/spaces/RsGoksel/turkish-asr-benchmark-stt}; data
  store \url{https://huggingface.co/datasets/RsGoksel/turkspeak-bench-store},
  2026, publicly released 2026-07-21; accessed 2026-07-29.

\bibitem[Thiemann et~al.(2013)Thiemann, Ito, and Vincent]{thiemann2013demand}
J.~Thiemann, N.~Ito, and E.~Vincent, ``The diverse environments multi-channel
  acoustic noise database ({DEMAND}): A database of multichannel environmental
  noise recordings,'' in \emph{Proceedings of Meetings on Acoustics (ICA
  2013)}, vol.~19, 2013, p. 035081.

\end{thebibliography}

\end{document}